\documentclass[10pt,journal,compsoc]{IEEEtran}

\usepackage{times}
\usepackage{epsfig}
\usepackage{graphicx}
\usepackage{amsmath}
\usepackage{amssymb}

\usepackage{mathtools}
\usepackage{bbm}
\usepackage{multirow}
\usepackage{booktabs} %

\usepackage{color}

\usepackage{float} 

\usepackage{algorithm}
\usepackage{algpseudocode}

\usepackage{multirow}
\usepackage{bm}
\usepackage{subcaption}

\usepackage{paralist}
\let\enumerate\compactenum
\let\endenumerate\endcompactenum

\usepackage{algorithm}
\usepackage{listings}
\usepackage[table]{xcolor}
\usepackage{booktabs} %
\usepackage{colortbl}
\usepackage{glossaries}
\usepackage{multirow}
\usepackage[font=small]{caption}
\usepackage{url}

\usepackage{pifont}%
\usepackage{xspace}
\usepackage{placeins}
\usepackage{tikz}
\usepackage{comment}
\usepackage{xspace}
\usepackage{amsmath,amssymb} %
\usepackage{color}
\usepackage{subcaption}

\ifCLASSOPTIONcompsoc
  \usepackage[nocompress]{cite}
\else
  \usepackage{cite}
\fi

\ifCLASSINFOpdf
\else
\fi

\begin{document}
\title{Introspective Deep Metric Learning}

\author{Chengkun~Wang*, Wenzhao~Zheng*, Zheng Zhu, \\
        Jie~Zhou,~\IEEEmembership{Senior Member,~IEEE,}
        and Jiwen~Lu,~\IEEEmembership{Senior Member,~IEEE}%
\IEEEcompsocitemizethanks{\IEEEcompsocthanksitem Chengkun Wang and Wenzhao Zheng contribute equally to this paper. 
Chengkun Wang, Wenzhao Zheng, Jie Zhou, and Jiwen Lu are with the State Key Lab of Intelligent Technologies and Systems, Beijing National Research Center for Information Science and Technology (BNRist), and the Department of Automation, Tsinghua University, Beijing, 100084, China. 
Zheng Zhu is with PhiGent Robotics, Beijing, 100084, China.
Email: wck20@mails.tsinghua.edu.cn; zhengwz18@mails.tsinghua.edu.cn; zhengzhu@ieee.org; jzhou@tsinghua.edu.cn; lujiwen@tsinghua.edu.cn.
}%
}

\markboth{IEEE TRANSACTIONS ON PATTERN ANALYSIS AND MACHINE INTELLIGENCE}%
{Shell \MakeLowercase{\textit{et al.}}: Bare Demo of IEEEtran.cls for Computer Society Journals}

\IEEEtitleabstractindextext{%
\begin{abstract}
This paper proposes an introspective deep metric learning (IDML) framework for uncertainty-aware comparisons of images. 
Conventional deep metric learning methods focus on learning a discriminative embedding to describe the semantic features of images, which ignore the existence of uncertainty in each image resulting from noise or semantic ambiguity. 
Training without awareness of these uncertainties causes the model to overfit the annotated labels during training and produce unsatisfactory judgments during inference.
Motivated by this, we argue that a good similarity model should consider the semantic discrepancies with awareness of the uncertainty to better deal with ambiguous images for more robust training.
To achieve this, we propose to represent an image using not only a semantic embedding but also an accompanying uncertainty embedding, which describes the semantic characteristics and ambiguity of an image, respectively.
We further propose an introspective similarity metric to make similarity judgments between images considering both their semantic differences and ambiguities. 
The gradient analysis of the proposed metric shows that it enables the model to learn at an adaptive and slower pace to deal with the uncertainty during training.
The proposed IDML framework improves the performance of deep metric learning through uncertainty modeling and attains state-of-the-art results on the widely used CUB-200-2011, Cars196, and Stanford Online Products datasets for image retrieval and clustering.
We further provide an in-depth analysis of our framework to demonstrate the effectiveness and reliability of IDML.
Code: \url{https://github.com/wzzheng/IDML}.

\end{abstract} 

\begin{IEEEkeywords}
Deep Metric Learning, Image Retrieval, Uncertainty-Aware Similarity Judgments.
\end{IEEEkeywords}}

\maketitle

\IEEEdisplaynontitleabstractindextext

\IEEEpeerreviewmaketitle

\newcommand{\tablevspace}{\vspace{-6mm}} %
\newcommand{\figvspace}{\vspace{-4mm}} %

\section{Introduction}
Learning an effective metric to measure the similarity between data is a long-standing problem in computer vision, which serves as a fundamental step in various downstream tasks, such as face recognition~\cite{schroff2015facenet,guo2020density,yang2019learning}, image retrieval~\cite{babenko2014neural,song2016deep,zheng2021deep} and image classification~\cite{deng2019arcface,shi2019probabilistic}. 
The general objective of metric learning is to reduce the distances between positive pairs and enlarge the distances between negative pairs, which has recently powered the rapid developments for both supervised learning~\cite{wang2020cross,zhu2020imbalance,wang2023deep} and unsupervised learning~\cite{he2020momentum,chen2021exploring,wang2022opera}. 

\begin{figure}[!t]
\centering
\includegraphics[width=0.495\textwidth]{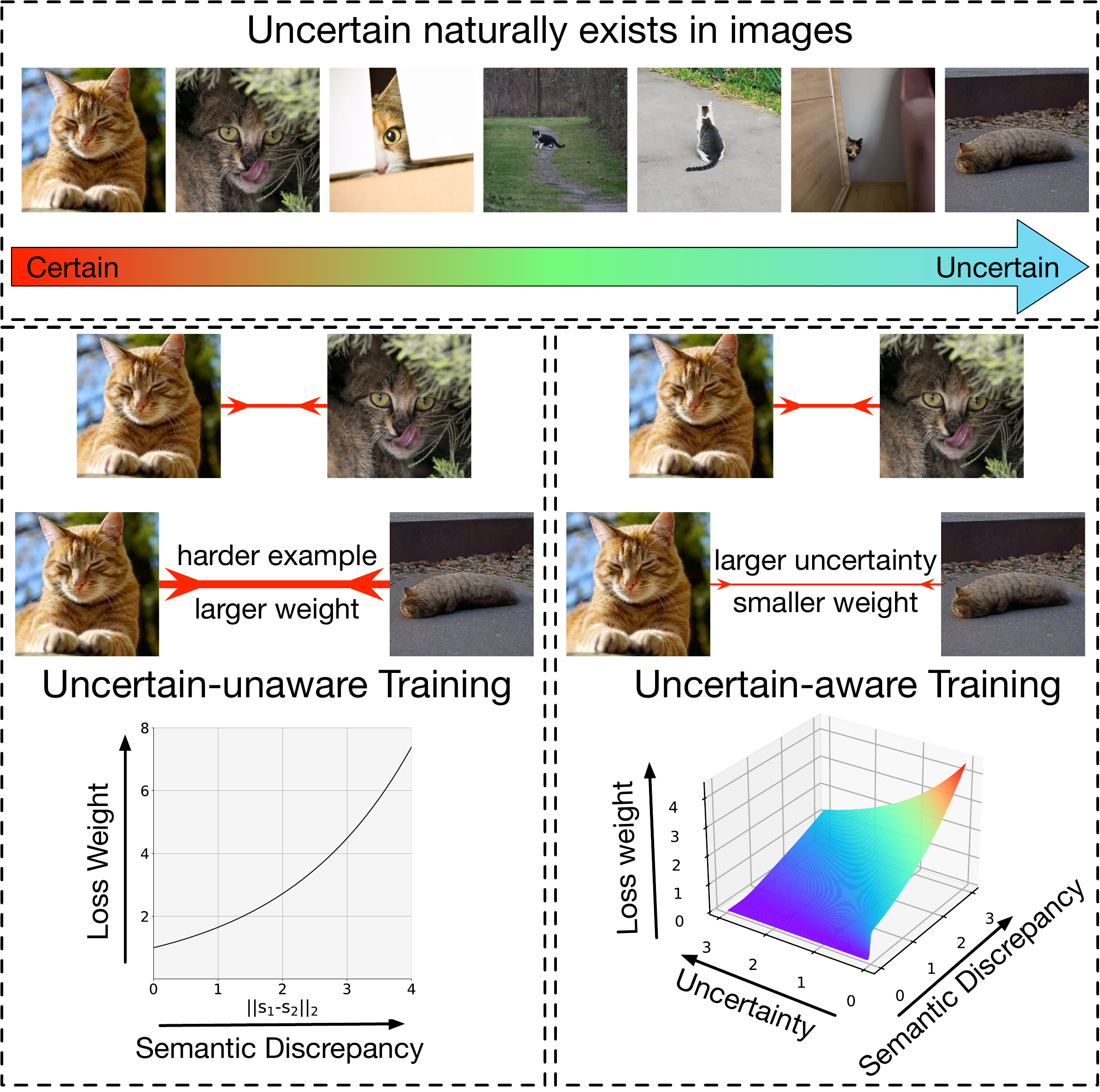}
\caption{
The motivation of the proposed IDML framework. 
Uncertainty naturally exists in images due to occlusion, scale, pose, low resolution, or semantic ambiguity. 
Most existing deep metric learning methods only consider the semantic distance of an image pair and determine its loss weight solely based on the semantic discrepancy.
However, we argue that the model should focus more on certain samples and put less weight on the more uncertain ones.
Our IDML achieves this by using both a semantic embedding and an uncertainty embedding to describe an image and employing an introspective similarity metric to compute an uncertainty-aware distance.
} 
\label{fig:motivation}
\figvspace
\end{figure}

Generally, deep metric learning (DML) employs deep neural networks~\cite{simonyan2014very,szegedy2015going,he2016deep} to map an image to a discriminative embedding space~\cite{zhang2017learning}.
Most methods represent images using a deterministic embedding which only describes the characteristic features~\cite{wu2017sampling,schroff2015facenet}.
Nevertheless, when asked to classify a certain image, humans are able to additionally provide the corresponding uncertainty as well as the semantic features of the image since an image might be ambiguous.
Motivated by this, researchers have proposed a variety of probabilistic embedding methods using distributions to model images in the embedding space~\cite{oh2018modeling,zhang2021point,sun2020view,chun2021probabilistic}. 
They typically use KL-divergence~\cite{hershey2007approximating} or Monte-Carlo-sampling-based~\cite{oh2018modeling} distances to measure the similarity between images.
They regard the variance of the distributions as the uncertainty measure of images, yet they still use these uncertain samples for similarity calculation and gradient update.
Specifically, though the variance affects the distribution discrepancy, a larger variance of an image does not necessarily blur its differences from other images.
Subsequently, existing probabilistic embedding methods only output an uncertainty score but the training process is still unaware of the uncertainty.
However, we argue that the model should focus more on certain samples and put less weight on the uncertain ones. 
This resembles humans who tend to ignore uncertain and uncredible information and occlude it from their learning process (\emph{e.g.}, experts engaged in the field of aerospace have never adopted uncertain technology), which enables humans to selectively and stably learn useful knowledge. 
Therefore, given a highly ambiguous image (\emph{e.g.}, a blurred ``6'' that is indistinguishable from other images in the MNIST~\cite{lecun1998gradient} dataset), we think it is more reasonable to weaken the semantic discrepancies and consider it similar to other images.

In this paper, we propose an introspective similarity metric to achieve this and further present an introspective deep metric learning (IDML) framework for image retrieval.
Different from existing methods, we represent an image using a semantic embedding to capture the semantic characteristics and further accompany it with an uncertainty embedding to model the uncertainty. 
An introspective similarity metric (ISM) then takes as input both embeddings and outputs an uncertainty-aware similarity score, which softens the semantic discrepancies by the degree of uncertainty.
Different from the conventional metric, the proposed introspective metric deems a pair of images similar if they are either semantically similar or ambiguous to judge.
It provides more flexibility to the training process to avoid the harmful influence of inaccurately labeled data, as illustrated in Fig.~\ref{fig:motivation}.
The proposed introspective similarity metric can alternatively enlarge the uncertainty level instead of rigidly enforcing the semantic constraint for an uncertain image.
We perform a gradient analysis of the loss function during the optimization process equipped with the proposed introspective similarity metric, which shows that IDML enables the model to learn at an uncertainty-aware pace.
The overall framework of the proposed IDML can be trained efficiently in an end-to-end manner and generally applied to existing deep metric learning methods.
Furthermore, we argue that the data uncertainty issue widely exists even for the general deep representation learning due to the commonly used data augmentation techniques.
We conduct experiments with different data augmentations including resizing, blurring, occlusion, and data mixing~\cite{zhang2018mixup,chou2020remix,verma2019manifold} and observe consistent improvements for our framework.
We perform extensive experiments on the CUB-200-2011, Cars196, and Stanford Online Products datasets for image retrieval and clustering, which shows that our framework generally improves the performance of existing deep metric learning methods by a large margin and attains state-of-the-art results.
We additionally provide an in-depth analysis of our framework including an ablation study of different components, effects of different augmentations, and qualitative analysis of the learned uncertainty.

We summarize the key contributions as follows:

\begin{enumerate}
	\item We propose an uncertainty-aware representation of images consisting of a semantic embedding and an uncertainty embedding. 
	    We further propose an introspective similarity metric to measure the similarity considering both semantic discrepancies and uncertainty levels.
	\item We propose an end-to-end introspective deep metric learning framework which can be generally applied to various deep metric learning methods.
		Our framework enables the model to learn at an uncertainty-adaptive pace.
	\item We conduct extensive experiments on image retrieval and observe the state-of-the-art performance of the proposed framework.
		We also provide an in-depth analysis of the working mechanism of our framework quantitively and qualitatively.
\end{enumerate}

\section{Related Work}
In this section, we present the related research in deep metric learning, sample weighting strategy, and uncertainty modeling.

\subsection{Deep Metric Learning}
Deep metric learning aims to construct an effective embedding space to measure the similarity between images, where the general objective is to decrease intraclass distances and increase interclass distances. 
Most existing methods~\cite{cakir2019deep,do2019theoretically,wang2014learning,yu2019deep} employ a discriminative loss to learn the image embeddings.
Instance-based methods directly restrict the distances between sample embeddings~\cite{hu2014discriminative,song2016deep,wang2017deep,wang2019multi,lu2017discriminative}. 
For example, the commonly used contrastive loss~\cite{hu2014discriminative} pulls embeddings from the same class as close as possible while maintaining a fixed margin between embeddings from different classes. 
Lu~\emph{et al.}~\cite{lu2017discriminative} proposed a distance ranking within the triplets and maintained a margin between positive pairs and negative pairs.
Wang~\emph{et al.}~\cite{wang2019multi} further formulated three types of similarities between embeddings and proposed a multi-similarity loss to restrict them. 
Proxy-based methods represent each class using proxy embeddings and instead constrain the similarities between sample embeddings and proxy embeddings~\cite{deng2019arcface,kim2020proxy,wang2018cosface,movshovitz2017no}.
The ProxyNCA loss~\cite{movshovitz2017no} generates a proxy for each class and simultaneously updates the proxies and sample embeddings. 
Kim~\emph{et al.}~\cite{kim2020proxy} further proposed to constrain both data-to-data relations and data-to-proxy relations.
Classification-based losses (\emph{e.g.}, CosFace~\cite{wang2018cosface} and ArcFace~\cite{deng2019arcface}) can also be regarded as proxy-based loss, where the class proxies are implicitly learned in the softmax classifier.

In addition to the design of loss functions, ensemble learning has been widely adopted deep metric learning research, which combines several weak learners for final predictions and has boosted the performances in various computer vision tasks~\cite{buckman2018sample,kurutach2018model,elghazel2015unsupervised,hong2008unsupervised}.
Generally, ensemble-based deep metric learning approaches improve the generalization ability by combining diverse embeddings supervised by relative tasks to encourage diversity of the learned representation~\cite{kim2018attention,sanakoyeu2019divide,milbich2020diva,yuan2017hard,sanakoyeu2019divide,xuan2018deep,kim2018attention}. For example, Yuan~\emph{et al.}~\cite{yuan2017hard} proposed to extract multi-level features from diverse layers of the backbone network to compose the weak learners. Additionally, Sanakoyeu~\emph{et al.}~\cite{sanakoyeu2019divide} divided the embedding space into sub-embeddings and assigned different samples to these sub-embeddings. However, the aforementioned deep metric learning methods treat samples without the consideration of sample uncertainty. Differently, we design an introspective similarity metric to consider the uncertainty and further incorporate them to make similarity judgments.

\subsection{Sample Weighting Strategy}
With the large amount of available samples for training, various methods have explored effective sample weighting strategies for more abundant distributions of training data~\cite{zhang2018mixup,yun2019cutmix,wu2017sampling,harwood2017smart,yuan2017hard,schroff2015facenet,duan2018deep,duan2019deep,roth2020pads,ge2018deep,yu2018correcting}. For example, hard negative mining~\cite{harwood2017smart,yuan2017hard,schroff2015facenet} selects challenging negative samples for more efficient learning of the metric, which improves the image retrieval performance as well as boosts the convergence speed.
Differently, Ge~\emph{et al.}~\cite{ge2018deep} focused on class representatives instead of instance-level samples for hard mining.
However, a distribution shift to the training data might be caused by hard negative sampling process thus Yu~\emph{et al.} corrected the selection bias by a distribution matching loss. 
In addition to hard mining, Wu~\emph{et al.}~\cite{wu2017sampling} further considered distance distributions to uniformly assign larger training weights for harder samples.
Roth~\emph{et al.}~\cite{roth2020pads} adaptively reweight samples with a reinforcement learning objective for triplet-based losses.
Furthermore, easy positive samples~\cite{xuan2020improved} have also been investigated to maintain the intra-class variance and thus improve the generalization ability on unseen classes.

To further alleviate the lack of informative training samples, recent works~\cite{duan2018deep,zheng2019hardness,zhao2018adversarial,lin2018deep} proposed to generate synthetic samples for training. 
For example, Duan~\emph{et al.} transformed easy samples into hard ones with an adversarially trained generator. 
Zheng~\emph{et al.} further proposed to synthesize hardness-aware adaptive samples.
Also, various data augmentation methods mix original images for better generalization~\cite{zhang2018mixup,yun2019cutmix}. 
The generated and augmented samples might lead to semantic ambiguity and label uncertainty.
However, existing methods usually reweight training samples according to their hardness without consideration of uncertainty.
Differently, IDML captures the uncertainty of training samples for more effective weighting.

\subsection{Uncertainty Modeling}
Uncertainty modeling is widely adopted in natural language processing to model the inherent hierarchies of words~\cite{vilnis2015word,nguyen2017mixture,neelakantan2014efficient}. 
Vilnis~\emph{et al.}~\cite{vilnis2015word} proposed the Gaussian formation in word embedding and Nguyen~\emph{et al.}~\cite{nguyen2017mixture} presented a mixture model to learn multi-sense word embeddings. 
Computer vision has also benefited from uncertainty modeling due to the natural uncertainty in images caused by factors such as occlusion and blur~\cite{kendall2017uncertainties,shaw2002signal}.
 Various methods have attempted to model the uncertainty for better robustness and generalization in face recognition~\cite{shi2019probabilistic,chang2020data}, point cloud segmentation~\cite{zhang2021point}, and age estimation~\cite{li2021learning}.
 In addition, uncertainty modeling is capable of handling several out-of-distribution problems, where the classes of testing samples might not be in the training data~\cite{guenais2020bacoun,li2022uncertainty,sun2022out,sehwag2021ssd}. For example, Guenais~\emph{et al.}~\cite{guenais2020bacoun} proposed the Bayesian classifiers with out-of-distribution uncertainty and Li~\emph{et al.}~\cite{li2022uncertainty} improved the out-of-distribution generalization ability of the network by modeling the uncertainty of domain shifts during training.

A prevailing method is to model each image as a statistical distribution and regard the variance as the uncertainty measure. 
For example, Oh~\emph{et al.}~\cite{oh2018modeling} employed Gaussian distributions to represent images and used Monte-Carlo sampling to sample several point embeddings from the distributions.
They then imposed a soft contrastive loss on the sampled embeddings to optimize the metric. 
Similar strategy has also been used in pose estimation~\cite{sun2020view}, cross-model retrieval~\cite{chun2021probabilistic} and unsupervised embedding learning~\cite{ye2020probabilistic}. 
However, they still use uncertain samples for similarity calculation and gradient update.
Therefore, a larger variance would not necessarily weaken the semantic discrepancies between samples.
Differently, we propose an introspective metric to measure the similarity between images, which tends to omit the semantic differences of two images given a large uncertainty level.
Our method bypasses the optimization of distributions and can further increase the robustness of the model.
\begin{figure}[t]
\centering
\includegraphics[width=0.495\textwidth]{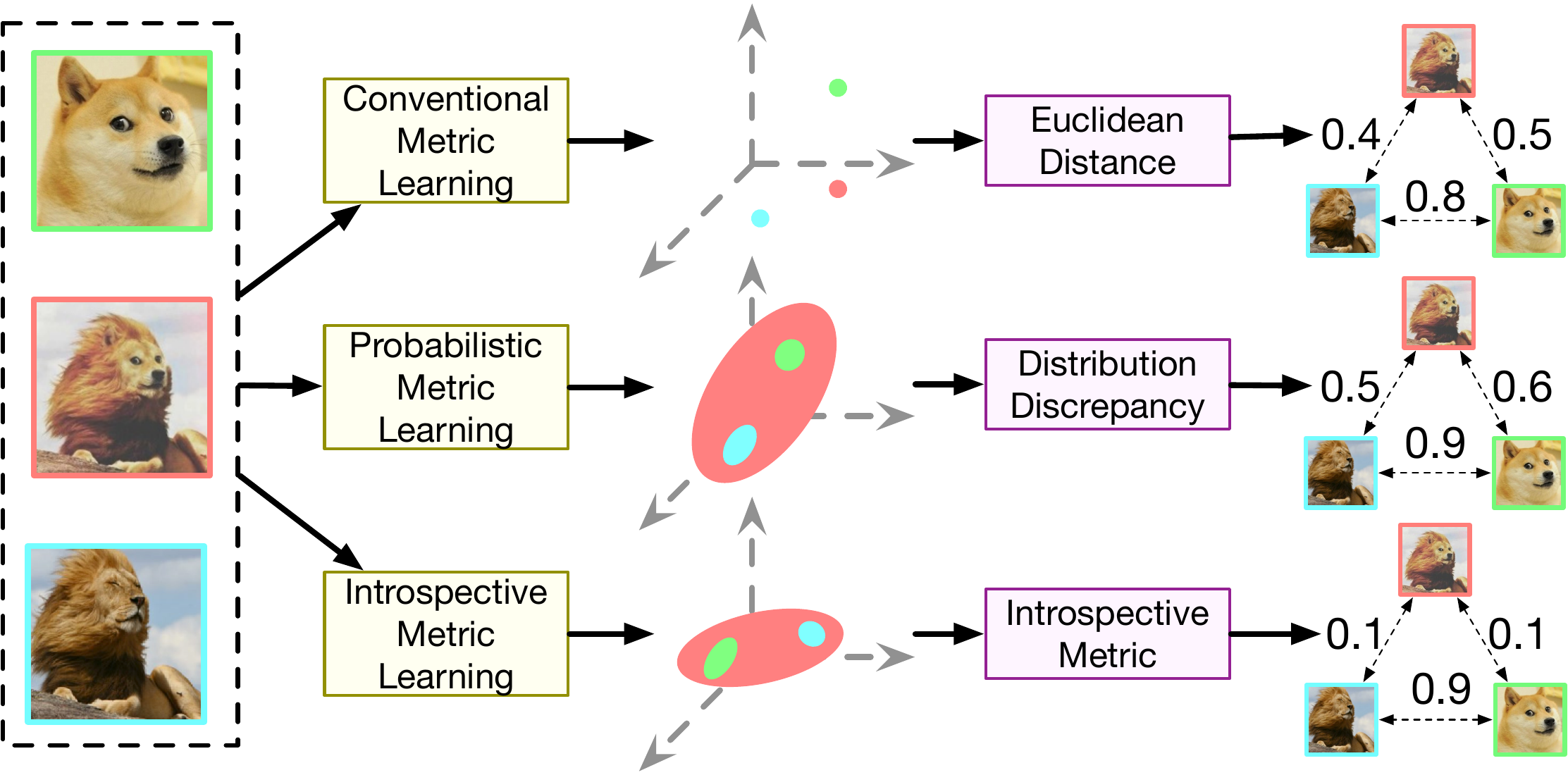}
\caption{
Comparisons between different similarity metrics. 
Conventional metric learning and probabilistic metric learning both produce a discriminative distance for a pair of images regardless of the uncertainty level.
Our introspective similarity metric weakens the semantic discrepancies for uncertain pairs.
} 
\label{fig:comparison}
\figvspace
\end{figure}

\section{Proposed Approach}
In this section, we first revisit existing deterministic and probabilistic deep metric learning methods and introduce the motivation of our framework.
We then present the introspective similarity metric and elaborate on the introspective deep metric learning framework.
Lastly, we conduct a gradient analysis to reveal the advantage of our framework and demonstrate how to apply our framework to existing deep metric learning methods.

\subsection{Motivation of an Uncertainty-Aware Metric}
Let $\mathbf{X}$ be an image set with $N$ training samples $\{\mathbf{x}_1, ..., \mathbf{x}_N\}$ and $\mathbf{L}$ be the ground truth label set $\{l_1, ..., l_N\}$. 
Deep metric learning methods aim at learning a mapping to transform each image $\mathbf{x}_i$ to an embedding space, where conventional methods~\cite{hu2014discriminative,movshovitz2017no} use a deterministic vector embedding $\mathbf{y}_i$ to represent an image.
They usually adopt the Euclidean distance as the distance measure:
\begin{eqnarray}\label{equ:euclidean}
D(\mathbf{x}_1, \mathbf{x}_2) = D_{E} (\mathbf{x}_1, \mathbf{x}_2) = ||\mathbf{y}_1 - \mathbf{y}_2||_2,
\end{eqnarray}
where $||\cdot||_2$ denotes the L2-norm.
They then impose various constraints on the pairwise distances, which generally enlarge interclass distances and reduce intraclass distances.

Conventional DML methods only encode semantic information in the embedding space, ignoring the possible uncertainty in images.
However, the semantic uncertainty ubiquitously exists due to low resolution, blur, occlusion, or semantic ambiguity, which motivates probabilistic embedding learning methods~\cite{oh2018modeling,chun2021probabilistic} to model images as statistical distributions $\mathbf{Y}$ in the embedding space. 
They further use the distribution variance $\mathbf{\sigma}$ to measure the uncertainty of the image in the embedding space.
One popular way is to employ a Gaussian distribution to describe an image~\cite{oh2018modeling}, \emph{i.e.}, $\mathbf{Y} \sim N(\mathbf{\mu}, \mathbf{\sigma})$, where they use a network to learn two vectors $\mathbf{\mu}$ and $\mathbf{\sigma}$ as the mean and variance of the distribution, respectively, assuming each dimension is independent.
They adopt distribution divergences~\cite{hershey2007approximating,yu2013kl} or Monte-Carlo-sampling-based~\cite{oh2018modeling} distances as the similarity metrics.
For instance, Hershey~\emph{et al.}~\cite{hershey2007approximating} used KL-divergence to measure the discrepancy of two Gaussian distributions $\mathbf{Y}_1$ and $ \mathbf{Y}_2$:
\begin{eqnarray}
 D_{KL} = 
 -\frac{1}{2} \sum_{k=1}^{d}[log\frac{\sigma_{1,k}^2}{\sigma_{2,k}^2} -  \frac{\sigma_{1,k}^2}{\sigma_{2,k}^2}  -  \frac{(\mu_{1,k}  -   \mu_{2,k})^2}{\sigma_{2,k}^2}   +   1],
\end{eqnarray}
where $d$ denotes the dimension of the Gaussian distributions, $\mathbf{Y}_1 \sim N(\mathbf{\mu}_1, \mathbf{\sigma}_1)$, and $\mathbf{Y}_2 \sim N(\mathbf{\mu}_2, \mathbf{\sigma}_2)$, and $\sigma_{1,k}$ denotes the $k$th component of $\mathbf{\sigma}_1$.

One can find that for two distributions with the same mean, their discrepancy solely depends on the ratio of the variance.
The discrepancy still varies greatly when the variance of one image is large, \emph{i.e.}, of large uncertainty. 
In other words, the distributional approaches still use these uncertain samples for similarity calculation and gradient update to a great extent.
However, we argue that a good similarity metric should weaken the semantic discrepancies for uncertain image pairs to allow confusion during training, which has been proven to be useful in knowledge distillation~\cite{hinton2015distilling}. 
We consider it reasonable to take the learned parameter as the sample uncertainty supposing the parameter can reduce the gradient of the training target for the sample.
This avoids the false pulling of ambiguous pairs to improve the generalization of the learned model.
Fig.~\ref{fig:comparison} presents the comparisons between different metrics.

\begin{figure}[t]
\centering
\includegraphics[width=0.495\textwidth]{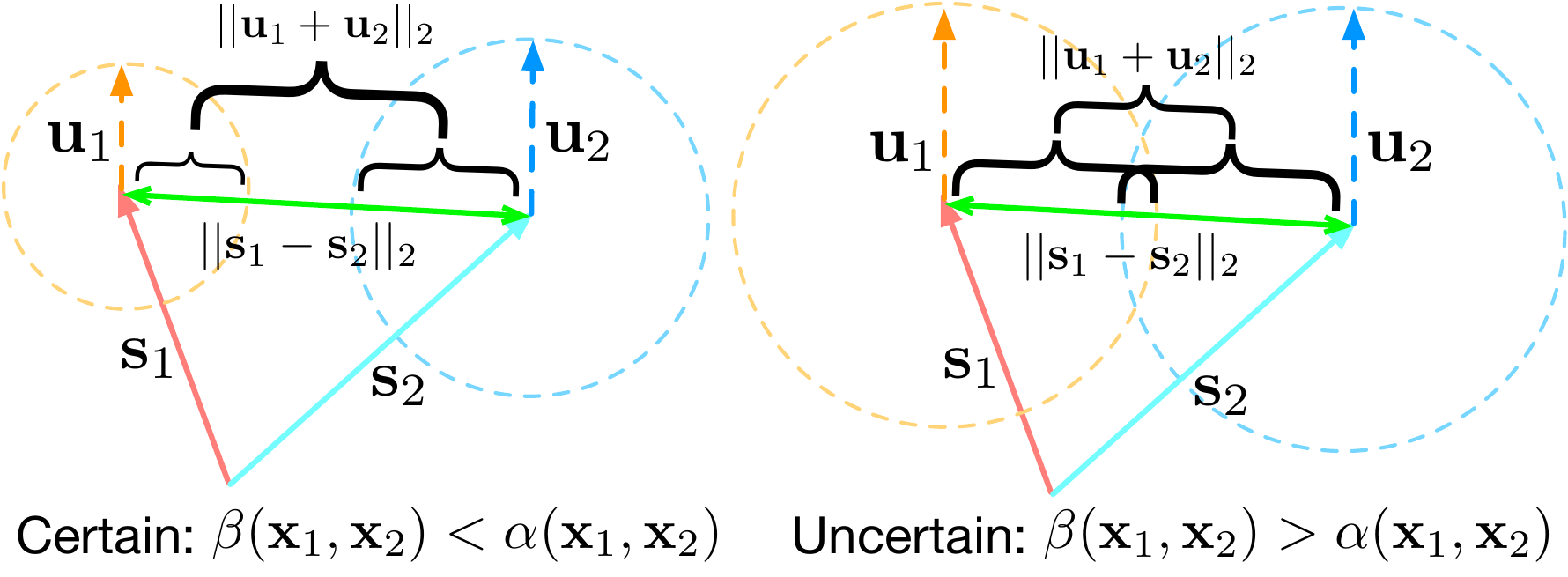}
\caption{
Illustration of the proposed uncertain-aware comparison of images.
We consider both the semantic discrepancy $\alpha(\mathbf{x}_1, \mathbf{x}_2) = ||\mathbf{s}_1 - \mathbf{s}_2 ||_2$ and the uncertainty level $\beta(\mathbf{x}_1, \mathbf{x}_2) = ||\mathbf{u}_1 + \mathbf{u}_2 ||_2$ to compute the similarity. 
We deem it uncertain to distinguish a pair when $\beta(\mathbf{x}_1, \mathbf{x}_2) > \alpha(\mathbf{x}_1, \mathbf{x}_2)$.
We only demonstrate the case when $\mathbf{u}_1$ and $\mathbf{u}_2$ align with each other for simplicity.
Practically, we use $||\mathbf{u}_1 + \mathbf{u}_2 ||_2$ instead of $||\mathbf{u}_1||_2 + ||\mathbf{u}_2 ||_2$ to facilitate more capacity.
} 
\label{fig:uncertain}
\figvspace
\end{figure}

\subsection{Introspective Similarity Metric}
To facilitate an uncertainty-aware similarity metric, we first need to model the uncertainty in images.
Different from existing probabilistic embedding learning methods to model images as distributions, we represent an image using a semantic embedding $\mathbf{s}$ and an uncertainty embedding $\mathbf{u}$, \emph{i.e.}, $\mathbf{y}_{IN} = \{ \mathbf{s}, \mathbf{u} \}$.
The semantic embedding $\mathbf{s}$ describes the semantic characteristics of an image while the uncertainty embedding $\mathbf{u}$ models the ambiguity. 

For comparing two images $\mathbf{x}_1$ and $\mathbf{x}_2$, we define the semantic distance as $\alpha(\mathbf{x}_1, \mathbf{x}_2) = ||\mathbf{s}_1 - \mathbf{s}_2 ||_2$ similar to conventional DML methods but further compute a similarity uncertainty as $\beta(\mathbf{x}_1, \mathbf{x}_2) = ||\mathbf{u}_1 + \mathbf{u}_2 ||_2$.
Note that we add the vectors of the uncertainty embeddings before computing the norm instead of directly adding their norms.
The reason is that the uncertainty should depend on both concerning images.
For example, it might be difficult to differentiate a burred ``6'' image in the MNIST~\cite{lecun1998gradient} dataset from other ``0'' samples, but it can be affirmatively distinguished from the ``3'' images.

When determining the semantic similarity between two images, an introspective metric needs to consider both the semantic distance and the similarity uncertainty. 
And when not certain enough, the metric refuses to distinguish semantic discrepancies, as illustrated in Fig.~\ref{fig:uncertain}.
Formally, we consider it uncertain to identify the semantic discrepancy between $\mathbf{x}_1$ and $\mathbf{x}_2$ when:
\begin{eqnarray}\label{equ:reckless}
\beta(\mathbf{x}_1, \mathbf{x}_2) + \gamma \geq \alpha(\mathbf{x}_1, \mathbf{x}_2),
\end{eqnarray}
where $\gamma \geq 0$ is the introspective bias indicating the introspective degree of the metric.
The positive value of $\gamma$ represents that the metric is still suspicious even if the image representation model provides no uncertainty.
We then define a strict introspective similarity metric as:
\begin{eqnarray}
\Tilde{D}_{IN}(\mathbf{x}_1,\mathbf{x}_2)= \alpha(\mathbf{x}_1,\mathbf{x}_2) \cdot I(\alpha(\mathbf{x}_1, \mathbf{x}_2) - \beta(\mathbf{x}_1, \mathbf{x}_2) - \gamma), 
\end{eqnarray}
where $I(x) $ is an indicator function which outputs 1 if $x>0$ and 0 otherwise.

However, the use of an indicator function is too strict and hard to optimize during training. 
We instead compare the semantic distance and the similarity uncertainty to define the relative uncertainty of two images:
\begin{eqnarray} \label{equ:r_unc}
\widetilde{\beta}(\mathbf{x}_1, \mathbf{x}_2) = \frac{\beta(\mathbf{x}_1, \mathbf{x}_2) + \gamma}{\alpha(\mathbf{x}_1, \mathbf{x}_2)}.
\end{eqnarray}
Note that the relative uncertainty is constantly non-negative.
We then use it to soften the semantic discrepancy to obtain our introspective similarity metric (ISM):
\begin{eqnarray}\label{equ:ism}
{D}_{IN}(\mathbf{x}_1,\mathbf{x}_2)= \alpha(\mathbf{x}_1,\mathbf{x}_2) \cdot e^{(-\frac{1}{\tau} \  \widetilde{\beta}(\mathbf{x}_1, \mathbf{x}_2))},
\end{eqnarray}
$\tau >0$ is a hyperparameter to control the weakening degree. 
Note that the proposed ISM does not satisfy the triangular equation and thus is not a mathematically strict metric.
We follow existing works~\cite{yuan2019signal,nguyen2010cosine} to still refer to it as a metric.

\begin{figure}[t]
\centering
\includegraphics[width=0.495\textwidth]{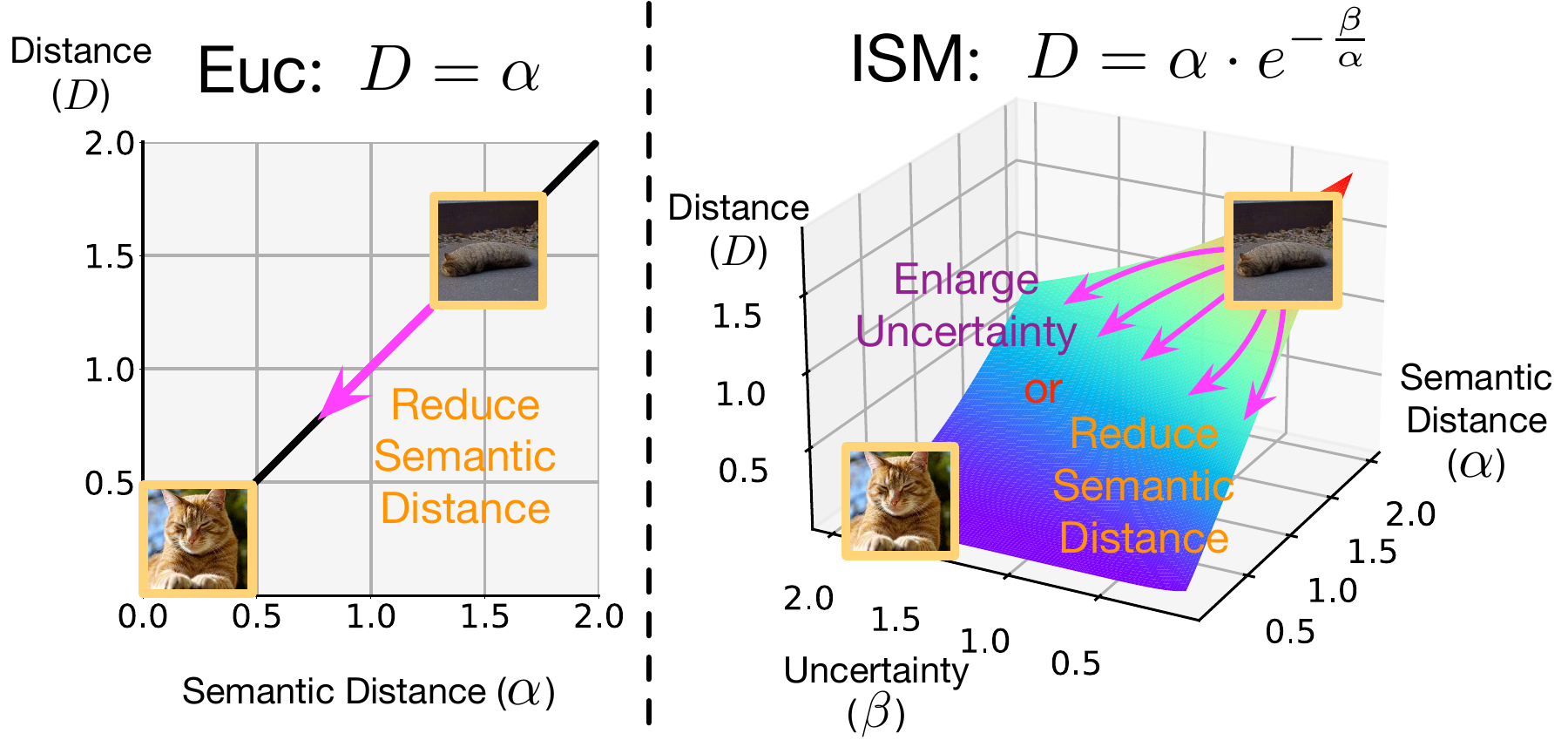}
\caption{
Illustration of the proposed ISM compared with the conventional Euclidean (Euc) metric.
For a semantically ambiguous image, conventional DML explicitly reduces its distance with other intraclass images unaware of the uncertainty.
Differently, the proposed introspective similarity metric provides an alternative way to enlarge the uncertainty level to allow confusion in the network.  
} 
\label{fig:ISM}
\figvspace
\end{figure}

Intuitively, the proposed introspectively metric considers both the semantic distance and the similarity uncertainty between two images to conduct the final semantic discrepancies.
It generally produces a smaller distance for two images due to the awareness of the uncertainty. 
Given two pairs of images with the same semantic distance, the introspective metric distinguishes better for the pair with a smaller similarity uncertainty.
Also, when the uncertainty of two images outweighs the semantic distance to a great extent, the introspective metric simply outputs a near-zero semantic distance avoiding unnecessary influence on the network. 
Fig.~\ref{fig:ISM} illustrates the effect of the proposed introspective similarity metric compared with the conventional Euclidean metric.
Our ISM provides an alternate way to enlarge the uncertainty for an ambiguous pair instead of rigidly altering their semantic distance.

\begin{figure*}[t]
\centering
\includegraphics[width=0.98\textwidth]{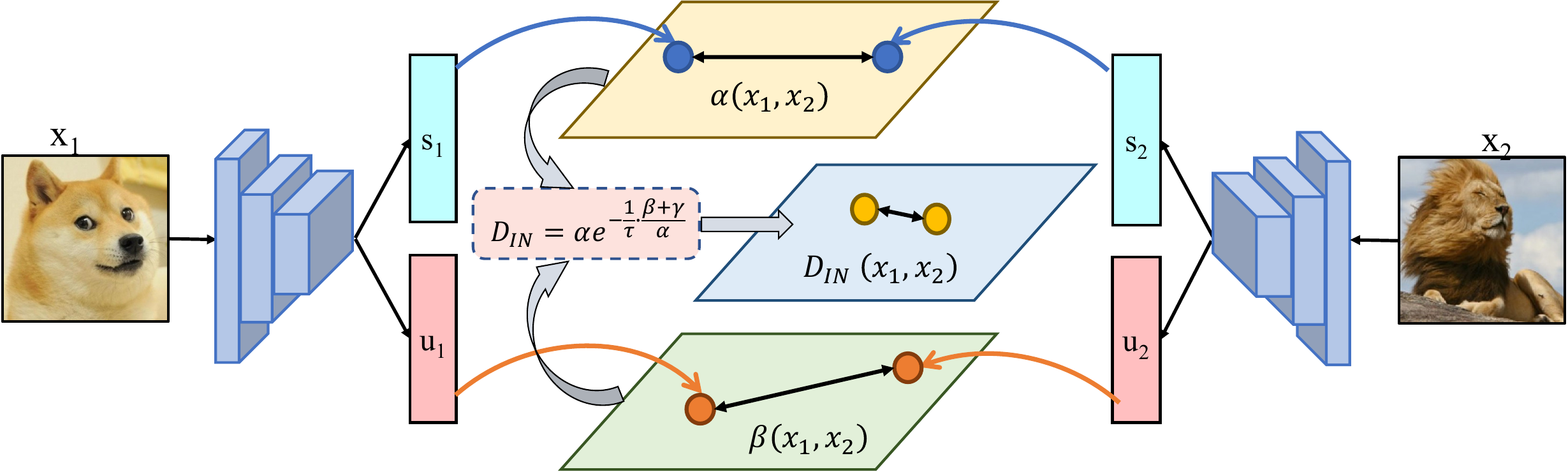}
\caption{
An illustration of the proposed IDML framework. 
We employ a convolutional neural network to represent each image by a semantic embedding and an uncertainty embedding. 
We then use the distance between semantic embeddings as the semantic discrepancy and add the uncertainty embeddings for uncertainty measure.
The introspective similarity metric then uses the uncertainty level to weaken the semantic discrepancy to make a discreet similarity judgment.
} 
\label{fig:overall}
\figvspace
\end{figure*}

\subsection{Introspective Deep Metric Learning}
In this subsection, we demonstrate how to apply the proposed introspective similarity metric to existing methods and present the overall framework of IDML, as illustrated in Fig.~\ref{fig:overall}.

Although data uncertainty ubiquitously exists in the original images, it is hard to accurately quantify and compare the uncertainty for each image.
On the other hand, data augmentation~\cite{zhang2018mixup,yun2019cutmix,duan2018deep} is a widely-used technique for training modern deep learning models.
They expand the training data to improve the generalization ability of the learned model, yet many of the augmented images (\emph{e.g.}, mixed, blurred, and partially occluded images) show larger semantic uncertainties compared to the original ones.
For example, one of the most effective data augmentation methods is Mixup~\cite{zhang2018mixup}, which randomly interpolates the pixel values of two images to generate a synthetic image with multiple concepts.
The generated images thus show characteristics of both the original images and are semantically uncertain to a great extent.
Therefore, we mainly accompany our framework with Mixup~\cite{zhang2018mixup} to demonstrate the advantage of our framework to deal with data with large uncertainty, though we find simply using ISM also improves the performance (Section \ref{analysis}).

Formally, Mixup~\cite{zhang2018mixup} mixes the original images $\mathbf{x}_1$ and $\mathbf{x}_2$ to obtain $\mathbf{x}_m=\lambda \cdot \mathbf{x}_1+(1-\lambda)\cdot \mathbf{x}_2$.
Different from the original method which combines the labels of two images by $l_m=\lambda \cdot l_1+(1-\lambda)\cdot l_2$, we treat $l_m$ as a set which simultaneously includes $l_1$ and $l_2$, noted as $l_m=\{l_1,l_2\}$. 
We define $l_i=l_j$ if $l_i \cap l_j \neq \emptyset$ and $l_i \neq l_j$ otherwise. 
In other words, for a mixed image of dog-lion, we think it belongs to both classes of dog and lion. Under such circumstances, we model the uncertainty of each image in a more reasonable manner, which essentially extracts the information from mixed samples.

We adopt a deep neural network to extract the feature embeddings of both original images and mixed images $\mathbf{y}_{IN}=f(\mathbf{x})=\{\mathbf{s},\mathbf{u}\}$, where $\mathbf{s}$ and $\mathbf{u}$ denote the semantic embedding and the uncertainty embedding of the image $\mathbf{x}$, respectively.
The distance (or similarity) computation of a pair of samples then follows the proposed introspective similarity metric \eqref{equ:ism}.
Our IDML can be generally applied to various methods with different loss functions and sampling strategies.
For a training objective $J(\mathbf{y}, \mathbf{L}; D)$, where $D$ can be the Euclidean distance, we simply substitute the embedding $\mathbf{y}$ and the metric $D$ with the proposed introspective embedding $\mathbf{y}_{IN}$ and the corresponding metric $D_{IN}$ as the objective $J(\mathbf{y}_{IN}, \mathbf{L}; D_{IN})$ of the proposed IDML framework.
We detail the specific objective with different methods in Section~\ref{implementation}.

For inference, we directly use the Euclidean distance between semantic embeddings as the similarity metric and can optionally use the uncertainty embedding to indicate the uncertainty level.
Therefore, the proposed IDML framework introduces negligible additional computation load compared to the original method.

\subsection{Applications of IDML to Various Methods} \label{implementation}
The proposed ISM is generally compatible with a variety of loss formulations and sampling strategies. 
Our framework can be readily applied to existing methods and can be very easily implemented with only a few additional lines of code.
We provide the PyTorch-like pseudocode of IDML in Algorithm~\ref{alg:idml}.

Most pair-based losses compute distances between samples, so we can directly employ the proposed ISM \eqref{equ:ism}.
However, many proxy-based losses such as the softmax loss~\cite{deng2019arcface,sun2020circle} usually compute the cosine similarity $C(\mathbf{x}_i, \mathbf{p}_j)$ instead of the distance between an image $\mathbf{x}_i$ and a class-level representative $\mathbf{p}_j$ (\emph{i.e.} proxy). 
To accommodate this, we propose a similarity-based version of the proposed ISM:
\begin{eqnarray}
{C}_{IN}(\mathbf{x}_i, \mathbf{p}_j)=1  -  (1-C(\mathbf{x}_i, \mathbf{p}_j))\cdot e^{(-\frac{1}{\tau} \  \widetilde{\beta}(\mathbf{x}_i, \mathbf{p}_j))},
\end{eqnarray}
where $\widetilde{\beta}(\mathbf{x}_i, \mathbf{p}_j)$ denotes the relative uncertainty between the image and the representative.
We see that it similarly blurs the similarity discrepancy for a larger uncertainty.

\textbf{Contrastive Loss:}
The contrastive loss~\cite{hu2014discriminative} directly pulls closer positive samples and pushes away negative samples:
\begin{eqnarray}
J_{con}(\mathbf{y}_{IN},\mathbf{L}; D_{IN}) = \sum_{l_i=l_j}D_{IN}(\mathbf{x}_i,\mathbf{x}_j) \nonumber \\
+\sum_{l_i\neq l_j}[\delta-D_{IN}(\mathbf{x}_i,\mathbf{x}_j)]_+,
\end{eqnarray}
where $\delta$ is the margin.

\textbf{Margin Loss \& Distance Weighted Sampling:}
The margin loss~\cite{wu2017sampling} is similar to the contrastive and additionally stipulates a margin to positive pairs. 
It is often equipped with the distance-weighted sampling strategy~\cite{wu2017sampling} for more uniform sampling:
\begin{eqnarray}
&& J_{m}(\mathbf{y}_{IN},\mathbf{L}; D_{IN}) = \sum_{l_i = l_j} [{D}_{IN} (\mathbf{x}_i, \mathbf{x}_j) - \xi]_+  \nonumber \\ 
&& -\sum_{l_i \neq l_j} I(p({D}_{IN} (\mathbf{x}_i, \mathbf{x}_j))) [\omega - {D}_{IN} (\mathbf{x}_i, \mathbf{x}_j)]_+,
\end{eqnarray}
where ${D}_{IN} (\mathbf{x}_i, \mathbf{x}_j)$ follows \eqref{equ:ism}, the random variable $I(p)$ has a probability of $p$ to be 1 and 0 otherwise, $p(d)=\min(\phi,d^{2-n}[1-\frac{1}{4}d^2]^{\frac{3-n}{2}})$, $[\cdot]_+ = \max(\cdot,0)$, $\xi$ and $\omega$ are two pre-defined margins, and $\phi$ is a positive constant.

\textbf{Triplet Loss \& Semi-hard Negative Sampling:}
The triplet loss~\cite{schroff2015facenet} requires a distance ranking within triplets and maintains a margin between positive pairs and negative pairs. 
The IDML with the triplet loss can be formulated as follows:
\begin{eqnarray}
J_{t}(\mathbf{y}_{IN},\mathbf{L}; D_{IN})\! = \!\!\! \sum_{a,p,n}[D_{IN}(\mathbf{x}_{a},\mathbf{x}_{p}) \! - \!D_{IN}(\mathbf{
x}_{a},\mathbf{x}_{n}) \! + \! \delta]_+,
\end{eqnarray}
where $D_{IN}(\cdot)$ denotes our introspective similarity metric, $\{a,p,n\}$ denote the indices of all the possible triplets, $[\cdot]_+ = \max(\cdot,0)$, and $\delta$ is a pre-defined margin. 
Furthermore, we employ the semi-hard negative sampling strategy~\cite{schroff2015facenet} to select challenging samples while avoiding noisy ones to boost training. 
Given an anchor $\mathbf{x}_a$ and a positive sample $\mathbf{x}_p$, we select the negative sample $\mathbf{x}_n$ which satisfies the following constraint:
\begin{eqnarray}
\mathbf{n}_{\mathbf{a},\mathbf{p}}^*=\mathop{\arg\min}\limits_{n:D_{IN}(\mathbf{x}_{\mathbf{a}},\mathbf{x}_{\mathbf{p}})<D_{IN}(\mathbf{x}_{\mathbf{a}},\mathbf{x}_{\mathbf{n}})}D_{IN}(\mathbf{x}_{\mathbf{a}},\mathbf{x}_{\mathbf{n}}).
\end{eqnarray}

\textbf{Multi-Similarity Loss:}
The multi-similarity (MS) loss~\cite{wang2019multi} uses the cosine similarity to measure the relations between samples. 
We first define a modified introspective similarity as follows:
\begin{eqnarray}\scriptsize
C_{IN}^*(\mathbf{x}_i,\mathbf{x}_j)\!\!=\!\!
\begin{cases}
C_{IN}(\mathbf{x}_i,\mathbf{x}_j),  \!\! & \!\! C_{IN}(\mathbf{x}_i,\mathbf{x}_j)\!>\!\mathop{\min}\limits_{l_k=l_i}C_{IN}(\mathbf{x}_i,\mathbf{x}_k)-\epsilon,\\
C_{IN}(\mathbf{x}_i,\mathbf{x}_j),\!\! & \!\! C_{IN}(\mathbf{x}_i,\mathbf{x}_j)\!<\!\mathop{\max}\limits_{l_k\neq l_i}C_{IN}(\mathbf{x}_i,\mathbf{x}_k)-\epsilon,\\ 
0,& otherwise, 
\end{cases}
\end{eqnarray}
where $\epsilon$ is a hyperparameter.
We then instantiate the IDML framework with the MS loss as follows:
\begin{equation}
\begin{aligned}
J_{MS}(\mathbf{y}_{IN},\mathbf{L}; C_{IN}^*) \!\! &= \!\! \frac{1}{N} \! \sum_{i=1}^N\left\{\frac{1}{\alpha}log[1+\!\!\!\sum_{l_i=l_j}e^{-\alpha(C_{IN}^*(\mathbf{x}_i,\mathbf{x}_j)-\lambda)}] \right.\\ 
&\left.+\frac{1}{\beta}log[1+\sum_{l_i\neq l_j}e^{\beta(C_{IN}^*(\mathbf{x}_i,\mathbf{x}_j)-\lambda)}]
\right\},
\end{aligned}
\end{equation}
where $\alpha$, $\beta$, and $\lambda$ are hyperparameters.

\begin{algorithm}[t]  
   \caption{Pseudocode of IDML in a PyTorch-like style.}
   \label{alg:idml}
    \definecolor{codeblue}{rgb}{0.25,0.5,0.5}
    \lstset{
      basicstyle=\fontsize{7.2pt}{7.2pt}\ttfamily\bfseries,
      commentstyle=\fontsize{7.2pt}{7.2pt}\color{codeblue},
      keywordstyle=\fontsize{7.2pt}{7.2pt},
    }
\begin{lstlisting}[language=python]
# X_tr, X_te: the training set, the testing set
# L_tr, L_te: the ground truth label sets
# f: the original network
# f_s, f_u: the semantic layer and uncertainty layer
# norm: l2 norm
# gamma, tau: pre-defined hyper-parameters
# J: the loss function
# V: the evaluation function

#training:
model.train()
for ((x1, x2), (l1, l2)) in (X_tr, L_tr):
    z1 = f(x1); z2 = f(x2) # feature extraction    
    s1 = f_s(z1); s2 = f_s(z2) # semantic embeddings
    u1 = f_u(z1); u2 = f_u(z2) # uncertainty embeddings    
    alpha(x1, x2) = norm(s1 - s2)
    beta(x1, x2) = norm(u1 + u2)    
    beta_r(x1, x2) = (beta(x1, x2) + gamma)/alpha(x1, x2)
    D_in(x1, x2) = alpha(x1, x2)*exp(-beta_r(x1, x2)/tau)    
    loss = J(D_in(x1, x2), l1, l2)
    loss.backward()

# evaluation:
S = [] # embeddings
model.eval()
for (x, l) in (X_te, L_te):
    z = f(x) # feature extraction 
    s = f_s(z) # the semantic embedding
    S.append(s) # same as conventional methods

evaluation_results = V(S) # introduce no additional load
\end{lstlisting}
\end{algorithm}

\textbf{Softmax Loss:}
The softmax loss is a very commonly used loss function mainly for the classification task.
It can be seen as a proxy-based metric learning loss, which maximizes the similarity of a sample with the positive proxy and minimizes the added similarities with other proxies:
\begin{eqnarray}
J_{s}(\mathbf{y}_{IN},\mathbf{L}; C_{IN}) = \frac{1}{N}\sum_{i=1}^{N}(-log\frac{\sum_{l_{\mathbf{p}_j}=l_{i}} e^{C_{IN}(\mathbf{x}_i,\mathbf{p}_j)}}{\sum_{l_{\mathbf{p}_j}\neq l_{i}}e^{C_{IN}(\mathbf{x}_i,\mathbf{p}_j)}}),
\end{eqnarray}
where $l_{\mathbf{p}_j}$ is the category of the class representative $\mathbf{p}_j$.
$\mathbf{p}_j$ can be implemented by the $j$-th row vector of a linear classifier.

Therefore, our IDML framework can be naturally adopted to the image classification task by involving the proposed introspective similarity in the corresponding softmax loss.

\textbf{ProxyNCA Loss:}
Very similar to the softmax loss, the ProxyNCA loss~\cite{movshovitz2017no} optimizes the distances between a sample and all the proxies instead of the similarities:
\begin{eqnarray}
J_{p}(\mathbf{y}_{IN},\mathbf{L}; D_{IN})=\sum_{i}(-log\frac{\sum_{l_{\mathbf{p}}=l_{i}} e^{- D_{IN}(\mathbf{x}_i,\mathbf{p})}}{\sum_{l_{\mathbf{p}}\neq l_{i}} e^{- D_{IN}(\mathbf{x}_i,\mathbf{p})}}),
\end{eqnarray}
where $l_i$ denotes the corresponding label of $\mathbf{x}_i$.

\textbf{ProxyAnchor Loss:}
The ProxyAnchor loss~\cite{kim2020proxy} takes advantage of both sample-sample and sample-proxy relations with the cosine similarity to improve the discriminativeness of the embedding space. 
Our IDML framework can be implemented for the ProxyAnchor loss as follows:
\begin{eqnarray}
J_{pa}(\mathbf{y}_{IN},\mathbf{L}; C_{IN})\!\!\!\!\!&=& \!\!\!\!\!
\frac{1}{|\mathbf{P}^+|}\sum_{\mathbf{p}\in \mathbf{P}^+}log(1+\sum_{l_i=l_{\mathbf{p}}} e^{-\alpha(C_{IN}(\mathbf{x}_i,\mathbf{p})-\delta)}) \nonumber \\
&+& \!\!\!\!\! \frac{1}{|\mathbf{P}|}\sum_{\mathbf{p}\in \mathbf{P}}log(1+\sum_{l_i\neq l_{\mathbf{p}}} e^{\alpha(C_{IN}(\mathbf{x}_i,\mathbf{p})+\delta)}), 
\end{eqnarray}
where $\mathbf{P}$ denotes the set of all proxies, $\mathbf{P}^+$ denotes the set of positive proxies, $|\cdot|$ represents the size of the set, $\alpha>0$ is a scaling factor, and $\delta>0$ is a pre-defined margin.

\begin{figure}[t]
\centering
\subcaptionbox{Effect of $\alpha$.\label{subfig:gradient-alpha}}{
\includegraphics[width=0.14\textwidth]{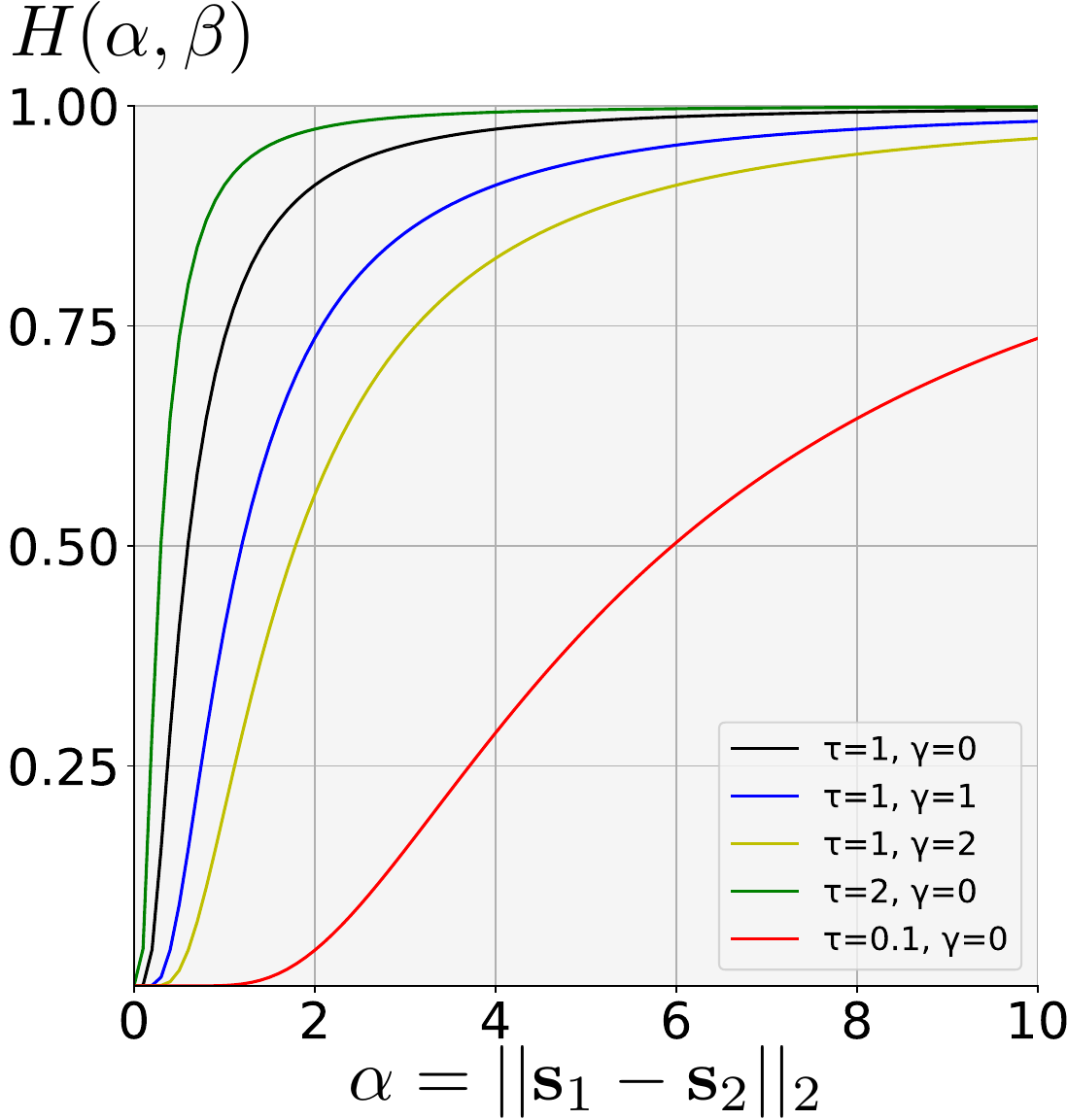}
}
\subcaptionbox{Effect of $\beta$.\label{subfig:gradient-beta}}{
\includegraphics[width=0.14\textwidth]{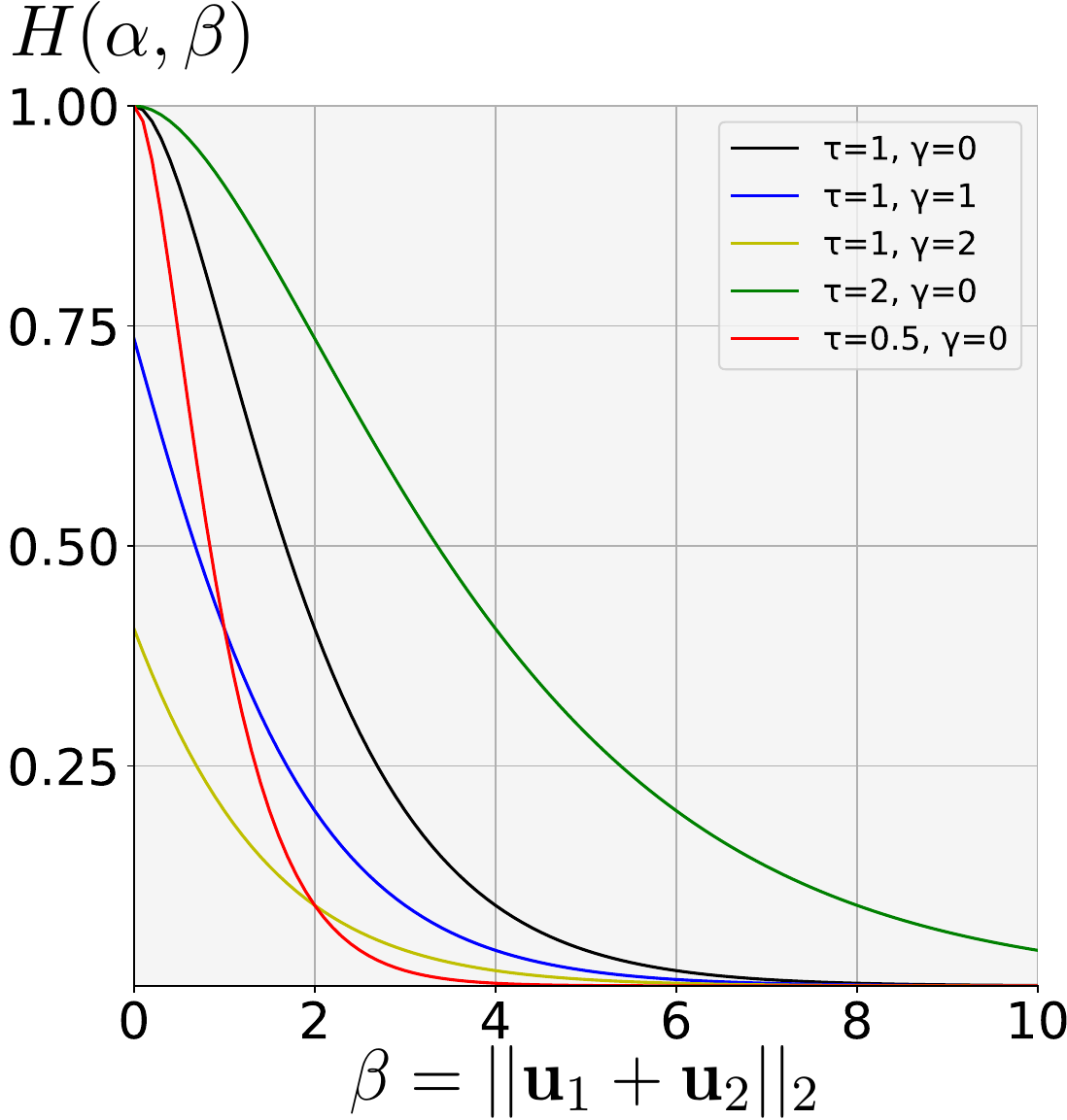}
}
\subcaptionbox{Effect of $\widetilde{\beta}$.\label{subfig:gradient-relative}}{
\includegraphics[width=0.14\textwidth]{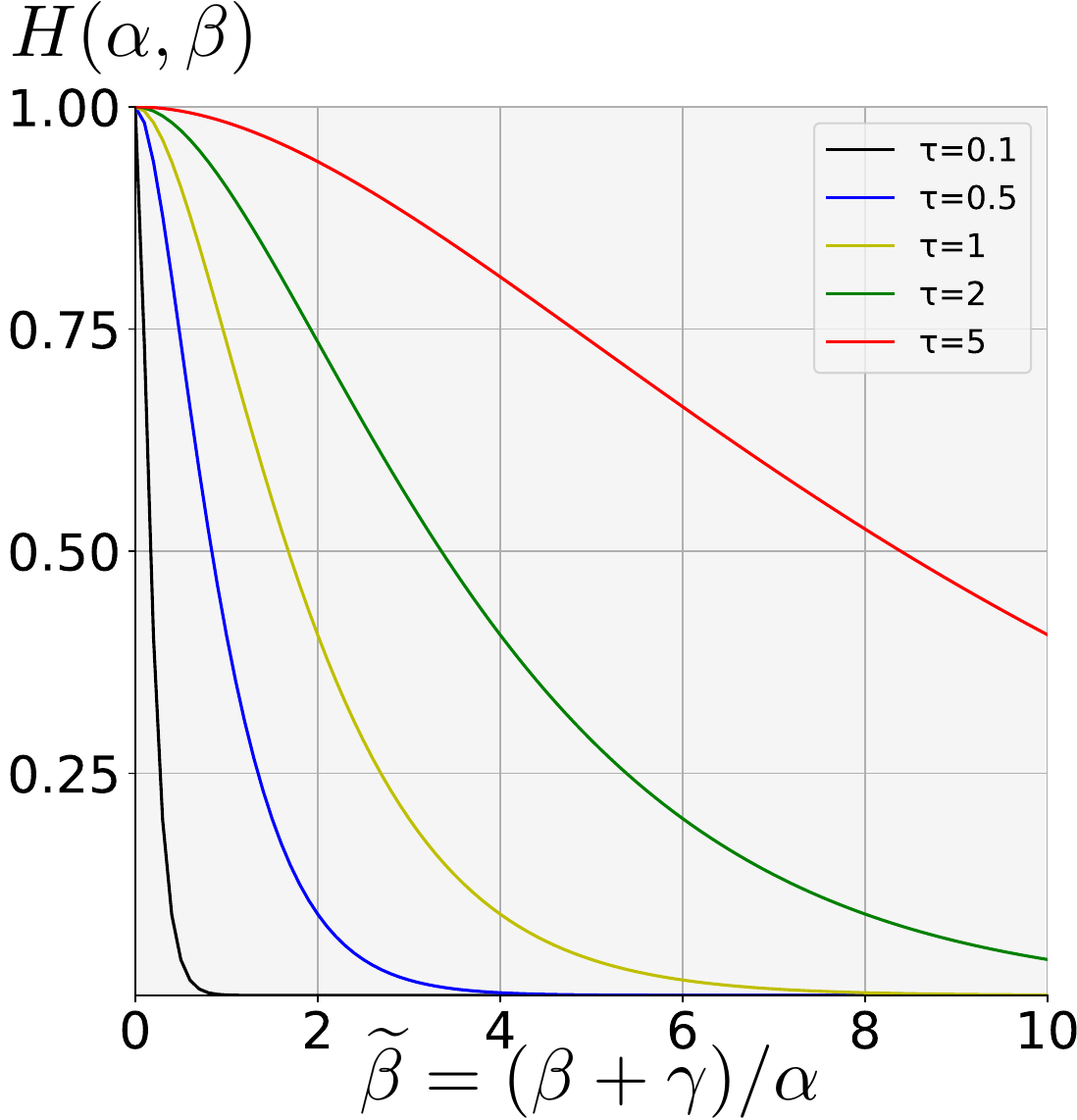}
}
\vspace{-3mm}
\caption{Influence of the (a) the semantic distance, (b) the uncertainty level, and (c) the relative uncertainty on the gradient weights.
With a weight generally less than 1, our ISM reduces the loss effects (\emph{i.e.}, gradients) on the network to avoid false training signals.
Furthermore, a larger similarity uncertainty and a smaller semantic distance indicate that the model tends not to differentiate two images, and the proposed ISM thus implicitly induces a smaller gradient.
}
\label{fig:gradient}
\figvspace
\end{figure}

\newcommand{\tablesize}{\small}
\newcommand{\arraysep}{\renewcommand\arraystretch{1.3}}

\subsection{Gradient Analysis}
We provide the gradient analysis to demonstrate the effect of our introspective similarity metric on the learning of the semantic embeddings.

We have claimed that our proposed similarity metric results in reduced semantic discrepancies to lower the influences of uncertain samples. 
These influences are measured by the magnitude of gradients on the model parameters.
For a conventional metric learning method $J^{E} = J(\mathbf{y}, \mathbf{L}; D_E)$ equipped with the Euclidean distance $D_E(\mathbf{x}_1, \mathbf{x}_2) = \alpha(\mathbf{x}_1, \mathbf{x}_2) = ||\mathbf{s}_1 - \mathbf{s}_2 ||_2$, we can decompose the loss gradient on the model parameters according to the chain rule as follows:
\begin{eqnarray} \label{equ:gradient_euc}
\frac{\partial J^{E}}{\partial W} = \frac{\partial J^{E}}{\partial \mathbf{s}}\cdot \frac{\partial \mathbf{s}}{\partial W} = \frac{\partial J^{E}}{\partial \alpha }\cdot \frac{\partial \alpha}{\partial \mathbf{s}}\cdot \frac{\partial \mathbf{s}}{\partial W},
\end{eqnarray}
where $W$ represents the model parameters that affect the semantic embedding $\mathbf{s}$. 
Similarly, for the IDML objective $J^{IN} = J(\mathbf{y}_{IN}, \mathbf{L}; D_{IN})$, we compute the loss gradient for our IDML framework as follows:
\begin{eqnarray} \label{equ:gradient_in}
\frac{\partial J^{IN}}{\partial W} 
\!=\! \frac{\partial J^{IN}}{\partial {D}_{IN}} \! \cdot \! \frac{\partial {D}_{IN}}{\partial \mathbf{s}} \! \cdot \! \frac{\partial \mathbf{s}}{\partial W} 
\! = \! \frac{\partial J^{IN}}{\partial {D}_{IN}} \! \cdot \! \frac{\partial {D}_{IN}}{\partial \alpha} \! \cdot \! \frac{\partial \alpha}{\partial \mathbf{s}} \! \cdot \! \frac{\partial \mathbf{s}}{\partial W}.
\end{eqnarray} 
In both \eqref{equ:gradient_euc} and \eqref{equ:gradient_in}, the partial term $\frac{\partial \mathbf{S}}{\partial W^t}$ is only relevant to the architecture of the backbone network.
As our IDML only substitutes the Euclidean distance $D_E$ in the original training objective with the proposed ISM $D_{IN}$, we have $\frac{\partial J^{E}}{\partial \mathbf{s}} = \frac{\partial J^{IN}}{\partial {D}_{IN}}$, which is determined by the form of the loss function.
Therefore, our IDML multiplies the conventional loss gradient with an additional term related to the uncertainty level $\beta(\mathbf{x}_1, \mathbf{x}_2)$:
\begin{eqnarray} \label{equ:gradient_relation}
\frac{\partial J^{IN}}{\partial W} = \frac{\partial J^{E}}{\partial W} \cdot  \frac{\partial {D}_{IN}}{\partial \alpha} = \frac{\partial J^{E}}{\partial W} \cdot  H (\alpha, \beta),
\end{eqnarray}
where 
\begin{eqnarray} \label{equ:gradient_effect}
H(\alpha, \beta) 
\coloneqq  \frac{\partial {D}_{IN}}{\partial \alpha} 
= \frac{\partial (\alpha \cdot e^{-\frac{1}{\tau} \frac{\beta + \gamma}{\alpha}})}{\partial \alpha}
= e^{- \frac{\widetilde{\beta}}{\tau} } \cdot (1 + \frac{\widetilde{\beta}}{\tau}),
\end{eqnarray}
where $\widetilde{\beta} = \frac{\beta + \gamma}{\alpha}$ is the relative uncertainty defined in \eqref{equ:r_unc}.

We plot the effect of the semantic distance, the uncertainty level, and the relative uncertainty on $H(\alpha, \beta)$ in Figs. \ref{subfig:gradient-alpha}, \ref{subfig:gradient-beta}, and \ref{subfig:gradient-relative}, respectively.
With the function $g(x)=e^{-x}\cdot(1+x)$ being monotonically decreasing when $x \geq 0$ and the relative uncertainty $\widetilde{\beta}$ being constantly non-negative, we see that $H(\alpha, \beta)$ has a maximum value of 1 when $\widetilde{\beta} = 0$ and decreases as the uncertainty level increases. 
This indicates that samples with larger uncertainties will result in smaller gradients compared to the original method, and thus impose fewer influences on the network.
Our method reduces the effects of semantically ambiguous images to avoid false training signals.
In particular, when $\widetilde{\beta} = 0$ (\emph{i.e.}, absolutely certain), we have $H(\alpha, \beta) = 1$ and $\frac{\partial J^{IN}}{\partial W} = \frac{\partial J^{IN}}{\partial W}$.
The proposed IDML framework then degenerates to the conventional method with the Euclidean distance.
Our framework is a generalization of conventional metric learning if we set the introspective bias $\gamma = 0$ and assign a zero uncertainty level to all the sample pairs.

\begin{table*}[t] \tablesize
\centering
\caption{Experimental results (\%) on the CUB-200-2011 and Cars196 datasets compared with state-of-the-art methods.  * denotes our reproduced results under the same settings.}
\label{tab:sota}
\vspace{-3pt}
\setlength\tabcolsep{3pt}
\arraysep
\begin{tabular}{lc|ccccc|ccccc}
\hline
 & & \multicolumn{5}{|c}{CUB-200-2011} & \multicolumn{5}{|c}{Cars196} \\
 \hline
Method & Setting & R@1 & R@2 & NMI & RP &M@R & R@1 & R@2 & NMI & RP & M@R \\
\hline
A-BIER~\cite{opitz2018deep} & 512G & 57.5 & 68.7 & - & - & - & 82.0 & 89.0 & - & - & - \\
ABE-8~\cite{kim2018attention} & 512G & 60.6 & 71.5 & - & - & - & 85.2 & 90.5 & - & - & - \\
Ranked~\cite{wang2019ranked} & 1536BN & 61.3 & 72.7 & 66.1 & - & - & 82.1 & 89.3 & 71.8 & - & - \\
DREML~\cite{xuan2018deep} & 9216R & 63.9 & 75.0 & 67.8 & - & - & 86.0 & 91.7 & 76.4 & - & - \\
SoftTriple~\cite{qian2019softtriple} & 512BN & 65.4 & 76.4 & 69.3 & - & - & 84.5 & 90.7 & 70.1 & - & - \\
D \& C~\cite{sanakoyeu2019divide} & 128R & 65.9 & 76.6 & 69.6 & - & - & 84.6 & 90.7 & 70.3 & - & - \\
MIC~\cite{roth2019mic} & 128R & 66.1 & 76.8 & 69.7 & - & - & 82.6 & 89.1  & 68.4 & - & - \\
RankMI~\cite{kemertas2020rankmi} & 128R & 66.7 & 77.2 & 71.3 & - & - & 83.3 & 89.8 & 69.4 & - & - \\
CircleLoss~\cite{sun2020circle} & 512R & 66.7 & 77.4 & - & - & - & 83.4 & 89.8 & - & - & - \\
PADS~\cite{roth2020pads} & 128BN & 67.3 & 78.0 & 69.9 & - & - & 83.5 & 89.7 & 68.8 & - & - \\
DIML~\cite{zhao2021towards} & 512R & 68.2 & - & - & 37.9 & 26.9 & 87.0 & - & - & 39.0 & 29.4 \\
DCML~\cite{zheng2021deep} & 512R & 68.4 & 77.9 & 71.8 & - & - & 85.2 & 91.8 & 73.9 & - & - \\
DRML~\cite{zheng2021deepr} & 512BN & 68.7 & 78.6 & 69.3 & - & - & 86.9 & 92.1 & 72.1 & - & - \\
ProxyNCA++~\cite{teh2020proxynca++}  & 512R & 69.0 & 79.8 & \color{red}73.9 & - & - & 86.5 & 92.5 & 73.8 & - & - \\ 
DiVA~\cite{milbich2020diva} & 512R & 69.2 & 79.3 & 71.4 & - & - & 87.6 & 92.9 & 72.2 & - & - \\
NIR~\cite{roth2022non} & 512R & \color{blue}{70.5} & \color{red}{80.6} & 72.5 & - & - & \color{blue}89.1 & \color{blue}93.4 & 75.0 & - & - \\
\hline
Triplet-SH*~\cite{schroff2015facenet} & 512R & 63.6$\pm$0.1 & 75.5$\pm$0.2 & 67.9$\pm$0.1 & 35.1$\pm$0.4 & 24.0$\pm$0.2 & 70.8$\pm$0.2 & 81.7$\pm$0.3 & 64.8$\pm$0.2 & 31.7$\pm$0.3 & 21.1$\pm$0.6 \\
IDML-TSH & 512R & \textbf{65.3$\pm$0.2} & \textbf{76.5$\pm$0.2} & \textbf{69.5$\pm$0.1} & \textbf{36.2$\pm$0.2} & \textbf{25.0$\pm$0.3} & \textbf{73.7$\pm$0.1} & \textbf{84.0$\pm$0.1} & \textbf{67.3$\pm$0.1} & \textbf{33.8$\pm$0.4} & \textbf{24.1$\pm$0.2} \\
\hline
ProxyNCA*~\cite{movshovitz2017no} & 512R & 64.6$\pm$0.3 & 75.6$\pm$0.3 & 69.1$\pm$0.4 & 35.5$\pm$0.3 & 24.7$\pm$0.6 & 82.6$\pm$0.2 & 89.0$\pm$0.1 & 66.4$\pm$0.2 & 33.5$\pm$0.3 & 23.5$\pm$0.2 \\
IDML-PN & 512R & \textbf{66.0$\pm$0.1} & \textbf{76.4$\pm$0.3} & \textbf{70.1$\pm$0.2} & \textbf{36.5$\pm$0.3} & \textbf{25.4$\pm$0.1} & \textbf{85.5$\pm$0.2} & \textbf{91.3$\pm$0.2} & \textbf{69.0$\pm$0.1} & \textbf{36.1$\pm$0.3} & \textbf{26.4$\pm$0.4} \\
\hline
FastAP*~\cite{cakir2019deep} & 512R & 65.1$\pm$0.5 & 75.4$\pm$0.4 & 68.5$\pm$0.4 & 35.9$\pm$0.9 & 24.1$\pm$0.6 & 81.6$\pm$0.3 & 88.5$\pm$0.4 & 68.8$\pm$0.3 & 35.1$\pm$0.5 & 25.2$\pm$0.6 \\
IDML-FAP & 512R & \textbf{66.4$\pm$0.3} & \textbf{76.4$\pm$0.2} & \textbf{69.7$\pm$0.3} & \textbf{36.7$\pm$0.4} & \textbf{25.5$\pm$0.4} & \textbf{83.9$\pm$0.4} & \textbf{89.9$\pm$0.5} & \textbf{71.9$\pm$0.3} & \textbf{36.5$\pm$0.7} & \textbf{26.7$\pm$0.5} \\
\hline
Contrastive*~\cite{hu2014discriminative} & 512R & 65.6$\pm0.1$ & 76.5$\pm0.1$ & 68.9$\pm$0.2 & 36.5$\pm$0.2 & 24.7$\pm$0.1 & 82.7$\pm$0.2 & 89.6$\pm$0.1 & 69.5$\pm$0.2 & 35.8$\pm$0.3 & 25.7$\pm$0.2 \\
IDML-Con & 512R & \textbf{67.2$\pm$0.0} & \textbf{77.6$\pm$0.1} & \textbf{71.3$\pm$0.1} & \textbf{37.5$\pm$0.2} & \textbf{25.7$\pm$0.2} & \textbf{85.5$\pm$0.1} & \textbf{91.5$\pm$0.2} & \textbf{72.5$\pm$0.1} & \textbf{38.8$\pm$0.1} & \textbf{29.0$\pm$0.3} \\
\hline
Margin-DW*~\cite{wu2017sampling} & 512R & 65.9$\pm$0.2 & 77.0$\pm$0.2 & 69.5$\pm$0.1 & 36.0$\pm$0.4 & 24.9$\pm$0.3 & 82.6$\pm$0.4 & 88.7$\pm$0.6 & 69.3$\pm$0.3 & 36.4$\pm$1.1 & 26.5$\pm$0.8 \\
IDML-MDW & 512R & \textbf{67.9$\pm$0.1} & \textbf{78.3$\pm$0.2} & \textbf{72.1$\pm$0.1} & \textbf{37.2$\pm$0.3} & \textbf{26.1$\pm$0.4} & \textbf{86.1$\pm$0.7} & \textbf{91.7$\pm$1.0} & \textbf{73.0$\pm$0.8} & \textbf{39.2$\pm$0.8} & \textbf{29.7$\pm$1.3} \\
\hline
Multi-Sim*~\cite{wang2019multi} & 512R & 67.3$\pm$0.3 & 78.2$\pm$0.2 & 72.7$\pm$0.3 & 36.6$\pm$0.2 & 25.5$\pm$0.4 & 83.3$\pm$0.1 & 90.9$\pm$0.3 & 72.2$\pm$0.4 & 37.4$\pm$0.2 & 27.4$\pm$0.3 \\
IDML-MS & 512R & \textbf{69.0$\pm$0.2} & \textbf{79.5$\pm$0.3} & \textbf{73.5$\pm$0.1} & \color{blue}\textbf{38.5$\pm$0.6} & \textbf{27.2$\pm$0.7} & \textbf{86.3$\pm$0.2} & \textbf{92.2$\pm$0.4} & \textbf{74.1$\pm$0.5} & \textbf{40.0$\pm$0.5} & \textbf{30.8$\pm$0.4} \\
\hline
ProxyAnchor*~\cite{kim2020proxy} & 512R & 69.0$\pm$0.1 & 79.4$\pm$0.1 & 72.3$\pm$0.1 & \color{blue}38.5$\pm$0.2 & \color{blue}27.5$\pm$0.3 & 87.3$\pm$0.0 & 92.7$\pm$0.1 & \color{blue}75.7$\pm$0.1 & \color{blue}40.9$\pm$0.2 & \color{blue}31.8$\pm$0.1 \\
IDML-PA & 512R & \color{red}\textbf{70.7$\pm$0.1} & \color{blue}\textbf{80.2$\pm$0.2} & \color{blue}\textbf{73.5$\pm$0.1} & \color{red}\textbf{39.3$\pm$0.4} & \color{red}\textbf{28.4$\pm$0.3} & \color{red}\textbf{90.6$\pm$0.1} & \color{red}\textbf{94.5$\pm$0.1} & \color{red}\textbf{76.9$\pm$0.1} & \color{red}\textbf{42.6$\pm$0.2} & \color{red}\textbf{33.8$\pm$0.4} \\
\hline
\end{tabular}
\tablevspace
\end{table*}

\begin{table}[t] \tablesize
\centering
\caption{Experimental results (\%) on the Stanford Online Products dataset compared with state-of-the-art methods.  * denotes our reproduced results under the same settings.}
\label{tab:sota_sop}
\vspace{-3pt}
\setlength\tabcolsep{1pt}
\arraysep
\begin{tabular}{lc|cccc}
\hline
 & & \multicolumn{4}{|c}{Stanford Online Products} \\
 \hline
Method & Setting & R@1 & NMI & RP & M@R \\
\hline
A-BIER~\cite{opitz2018deep} & 512G & 74.2 & - & - & -\\
ABE-8~\cite{kim2018attention} & 512G & 76.3 & - & - & -\\
Ranked~\cite{wang2019ranked} & 1536BN & 79.8 & 90.4 & - & -\\
DREML~\cite{xuan2018deep} & 9216R & - & - & - & -\\
SoftTriple~\cite{qian2019softtriple} & 512BN & 78.3 & 92.0 & - & -\\
D \& C~\cite{sanakoyeu2019divide} & 128R & 75.9 & 90.2 & - & -\\
MIC~\cite{roth2019mic} & 128R & 77.2 & 90.0 & - & -\\
RankMI~\cite{kemertas2020rankmi} & 128R & 74.3 & 90.5 & - & -\\
CircleLoss~\cite{sun2020circle} & 512R 78.3 & 90.5 & - & -\\
PADS~\cite{roth2020pads} & 128BN & 76.5 & 89.9 & - & -\\
DIML~\cite{zhao2021towards} & 512R & 79.3 & - & 46.4 & 43.2\\
DCML~\cite{zheng2021deep} & 512R & 79.8 & 90.8 & - & -\\
DRML~\cite{zheng2021deepr} & 512BN & 79.9 & 90.1 & - & -\\
ProxyNCA++~\cite{teh2020proxynca++}  & 512R & \color{blue}80.7 & - & - & -\\ 
DiVA~\cite{milbich2020diva} & 512R & 79.6 & 90.6 & - & - \\
NIR~\cite{roth2022non} & 512R & \color{blue}80.7 & 90.9 & - & - \\
\hline
Triplet-SH*~\cite{schroff2015facenet} & 512R & 76.5$\pm$0.2 & 89.7$\pm$0.1 & 51.3$\pm$0.1 & 48.4$\pm$0.2\\
IDML-TSH & 512R & \textbf{77.4$\pm$0.1} & \textbf{90.1$\pm$0.1} & \textbf{51.9$\pm$0.3} & \textbf{49.0$\pm$0.2}\\
\hline
ProxyNCA*~\cite{movshovitz2017no} & 512R & 77.0$\pm$0.2 & 89.5$\pm$0.2 & 51.9$\pm$0.4 & 49.0$\pm$0.2\\
IDML-PN & 512R & \textbf{78.3$\pm$0.2} & \textbf{89.9$\pm$0.3} & \textbf{53.0$\pm$0.2} & \textbf{49.9$\pm$0.2}\\
\hline
FastAP*~\cite{cakir2019deep} & 512R & 75.9$\pm$0.6 & 89.7$\pm$0.4 & 50.1$\pm$0.7 & 46.8$\pm$0.6 \\
IDML-FAP & 512R & \textbf{76.8$\pm$0.4} & \textbf{90.9$\pm$0.4} & \textbf{50.9$\pm$0.6} & \textbf{47.9$\pm$0.6}\\
\hline
Contrastive*~\cite{hu2014discriminative} & 512R & 76.4$\pm$0.1 & 88.9$\pm$0.1 & 50.9$\pm$0.1 & 47.9$\pm$0.2 \\
IDML-Con & 512R & \textbf{77.3$\pm$0.0} & \textbf{90.0$\pm$0.1} & \textbf{51.7$\pm$0.2} & \textbf{48.5$\pm$0.1} \\
\hline
Margin-DW*~\cite{wu2017sampling} & 512R & 78.5$\pm$0.3 & 90.1$\pm$0.2 & 53.4$\pm$0.3 & 50.2$\pm$0.3 \\
IDML-MDW & 512R & \textbf{79.4$\pm$0.2} & \textbf{91.0$\pm$0.1} & \color{blue}\textbf{53.7$\pm$0.3} & \textbf{50.4$\pm$0.2}\\
\hline
Multi-Sim*~\cite{wang2019multi} & 512R & 78.1$\pm$0.2 & 89.9$\pm$0.2 & 52.9$\pm$0.1 & 49.9$\pm$0.3\\
IDML-MS & 512R & \textbf{79.7$\pm$0.1} & \color{blue}\textbf{91.2$\pm$0.1} & \color{blue}\textbf{53.7$\pm$0.2} & \color{blue}\textbf{50.9$\pm$0.3}\\
\hline
ProxyAnchor*~\cite{kim2020proxy} & 512R & 79.5$\pm$0.1 & 91.0$\pm$0.0 & \color{blue}53.7$\pm$0.1 & 50.5$\pm$0.1 \\
IDML-PA & 512R & \color{red}\textbf{81.5$\pm$0.2} & \color{red}\textbf{92.3$\pm$0.1} & \color{red}\textbf{54.8$\pm$0.2} & \color{red}\textbf{51.3$\pm$0.3}\\
\hline
\end{tabular}
\tablevspace
\end{table}

\section{Experiments}

In this section, we conducted various experiments to evaluate the performance of our IDML framework on image retrieval.
We show that employing the proposed introspective similarity metric consistently improves the performance of existing deep metric learning and data mixing methods.
We also provide in-depth analyses of the effectiveness of our framework.

\subsection{Settings}
We evaluated our framework under the conventional deep metric learning setting~\cite{song2016deep,wang2019multi} and conducted experiments on three widely-used datasets: CUB-200-2011~\cite{wah2011caltech}, Cars196~\cite{krause20133d}, and Stanford Online Products~\cite{song2016deep}. 
We adopted the ImageNet~\cite{russakovsky2015imagenet} pretrained ResNet-50~\cite{he2016deep} as the backbone and two randomly initialized fully connected layers to obtain the semantic embedding and uncertainty embedding, respectively.
We set the embedding size to 512 for the main experiments.
The training images were first resized to $256\times256$ and then augmented with random cropping to $224\times224$ as well as random horizontal flipping with the probability of $50\%$. 
We employed Mixup~\cite{zhang2018mixup} for our framework to generate images with large uncertainty for training unless otherwise stated.
We fixed the batch size to $120$ and used AdamW optimizer with the learning rate of $10^{-5}$. 
We set $\gamma=2$ for the Cars196 dataset and $\gamma=0$ for the other datasets and fixed $\tau=5$ for all the datasets during training. 
We adopted the original similarity metric without our uncertainty embedding for testing and thus introducing no additional computational workload.
The reported evaluation metrics include Recall@Ks, normalized mutual information (NMI), R-Precision (RP), and Mean Average Precision at R (M@R).
See Musgrave~\emph{et al.}~\cite{musgrave2020metric} for more details. We conducted all experiments using the PyTorch~\cite{paszke2019pytorch} library.

\subsection{Dataset}
For image retrieval and clustering, we followed existing DML methods~\cite{song2016deep, zheng2019hardness, wang2019multi, kim2020proxy} to conduct experiments on the CUB-200-2011~\cite{wah2011caltech}, Cars196~\cite{krause20133d}, and Stanford Online Products~\cite{song2016deep}. 
CUB-200-2011 contains 11,788 images of 200 bird species. 
We used the first 100 species with 5,864 images for training and the rest 100 species with 5,924 images are for testing. 
Cars196 includes 16,183 images of 196 car models. 
We used the first 96 classes with 8,054 images for training and the rest 98 classes with 8,131 images for testing. 
Stanford Online Products is relatively large and contains 120,053 images of 22,634 products. 
We used the first 11,318 products with 59,551 images in the training set and the rest of 11,318 products with 60,502 images for testing.

\subsection{Main Results}
We evaluated the proposed IDML framework under the conventional deep metric learning setting and compared it with state-of-the-art methods.
To demonstrate the versatility of our framework, we applied the introspective similarity metric to various loss functions, including the triplet loss with the semi-hard sampling (Triplet-SH)~\cite{schroff2015facenet}, the ProxyNCA loss~\cite{movshovitz2017no}, the FastAP loss~\cite{cakir2019deep}, the contrastive loss~\cite{hu2014discriminative}, the margin loss with the distance-weighed sampling (Margin-DW)~\cite{wu2017sampling}, the multi-similarity loss (Multi-Sim)~\cite{wang2019multi}, and the ProxyAnchor loss~\cite{kim2020proxy}. 

Table~\ref{tab:sota} and Table~\ref{tab:sota_sop} show the experimental results on the CUB-200-2011~\cite{wah2011caltech}, Cars196~\cite{krause20133d}, and Stanford Online Products~\cite{song2016deep} datasets. 
The n-G/BN/R denotes the model setting where n is the embedding size and G, BN, R represents GoogLeNet~\cite{szegedy2015going}, BN-Inception~\cite{ioffe2015batch} and ResNet-50~\cite{he2016deep}, respectively. 
The bold numbers highlight the improvement of our framework compared with the original method. We indicate the best results
using red colors and the second best results using blue colors.
We observe that our framework achieves a constant performance boost to all the associated deep metric learning methods. 
Furthermore, we attain state-of-the-art performances on all three datasets by applying our IDML framework to the ProxyAnchor loss (70.7$\%$, 90.6$\%$, and 81.5$\%$ at Recall@1 and 28.4$\%$, 33.8$\%$, and 51.3$\%$ at M@R on three datasets, respectively), which surpasses the original performance by $1.7\%$, $3.3\%$, and $2.0\%$ at Recall@1 and $0.9\%$, $2.0\%$, and $1.8\%$ at M@R on three datasets, respectively.
This is because the proposed similarity metric is aware of the data uncertainty in images so that the uncertainty information has been captured by the model and the uncertain samples only provide limited training signals.

\subsection{Analysis} \label{analysis}

We analyze the impact of various structures of the IDML framework and visualize several experimental results on image retrieval tasks.
We conducted extensive experiments under the conventional deep metric setting to analyze the effectiveness of our framework. 

\subsubsection{Comparison with the Distributional Modeling Method}
We have investigated a distribution modeling approach proposed by Scott~\emph{et al.}~\cite{scott2021mises}, denoted as vMF for simplicity. vMF serves as a distributional modeling strategy that obtains the mean unit vector $\mathbf{\mu}$ and the isotropic concentration $\kappa$ from each sample, where $\mathbf{\mu}$ represents the semantic vector of the sample while $\kappa$ denotes the corresponding uncertainty level. In the original paper, the authors proposed the pdf of vMF that was used to model the distributional feature of each image as well as to conduct the sampling strategy during training and testing, formulated as follows:
\begin{equation}
p(\mathbf{x};\mathbf{\mu},\kappa)=C_n(\kappa)exp(\kappa \mathbf{\mu}^T\mathbf{x})
\end{equation}
\begin{equation}
C_n(\kappa)=\frac{\kappa^{n/2-1}}{(2\pi)^{n/2}I_{n/2-1}(\kappa)}
\end{equation}
where $I_v$ represents the modified Bessel function of the first kind at order $v$, and a larger $\kappa$ denotes a smaller sample uncertainty.

Therefore, we compared vMF and our IDML in theory and through experiments, respectively: (1) vMF follows the general probabilistic embedding methods using distributions to model images in the embedding space. Although vMF utilizes an additional uncertainty parameter $\kappa$ but not the variance of the distribution to measure the uncertainties of samples, a smaller $\kappa$ (larger uncertainty) of an image does not necessarily blur its differences from other images, which means that uncertain samples still influence the training of the whole network. There, vMF can not avoid such imperfection similar to the probabilistic embedding approaches. (2) vMF represents the uncertainty of a sample with a single additional parameter, which definitely reduces the amount of parameters and is beneficial to the memory of the network. Nevertheless, vMF utilizes Monte-Carlo sampling to conduct the subsequent optimization process, which increases the running time during both training and testing. On the contrary, IDML saves processing time while increasing the number of parameters. (3) We have conducted experiments to utilize vMF to model each sample in our deep metric learning framework. Differently from the original paper, the class weights $\mathbf{w}_j$ are replaced by the predefined proxies $\mathbf{P}$. We have fixed other settings compared with IDML including the data augmentation strategies, the network architecture the loss function, and the chosen optimizer. We have tested the performances of vMF on three datasets, shown in Table~\ref{tab:vMF}. We observe that vMF improves the experimental results compared with the baseline method (70.7$\%$, 89.6$\%$, and 80.8$\%$ at Recall@1 on three datasets, respectively). However, our IDML surpasses vMF by 0.7$\%$, 1.0$\%$, and 0.7$\%$ at Recall@1 on all three datasets respectively, verifying that IDML might be more suitable for deep metric learning tasks. 

\subsubsection{Experimental Results of Out-of-distribution Detection}
We adopted CIFAR-10~\cite{krizhevsky2009learning} (C-10) and CIFAR-100~\cite{krizhevsky2009learning} (C-100) as in-distribution (ID) and out-of-distribution (OOD) following the baseline settings~\cite{sehwag2021ssd}. We combined the supervised learning recipe with vMF~\cite{scott2021mises} and our IDML framework. Specifically, we employed an additional uncertainty embedding and used ISM for similarity computation for IDML. We provided the performance boost and the uncertainties of training and testing samples in Table~\ref{tab:ood}. 
We can see that vMF indeed improves the original performances of OOD detection (AUROC: 91.2 when CIFAR-10 is ID and 57.6 when CIFAR-100 is ID).
However, we observe that IDML still outperforms vMF by 0.7 at AUROC when CIFAR-10 is ID and 1.0 at AUROC when CIFAR-100 is ID in this task.
In addition, the uncertainties of testing samples in IDML are obviously higher than those of training samples, verifying the reliability of the learned uncertainty embeddings.

\begin{table}[!t] \small
\centering
\caption{Comparisons between vMF and IDML on three datasets.}
\label{tab:vMF}
\vspace{-3pt}
\setlength\tabcolsep{7.5pt}
\arraysep
\begin{tabular}{l|l|cccc}
\hline
Dataset & Method & R@1 & NMI & RP & M@R\\
\hline
\multirow{3}{*}{CUB-200-2011}
& Baseline & 69.0 & 72.3 & 38.5 & 27.5\\
  & vMF & 70.0 & 72.8 & 38.9 & 28.0\\
  & IDML & \textbf{70.7} & \textbf{73.5} & \textbf{39.3} & \textbf{28.4}\\
\hline
\multirow{3}{*}{Cars196} 
  & Baseline & 87.3 & 75.7 & 30.9 & 31.8\\
  & vMF & 89.6 & 76.0 & 41.7 & 32.6\\
  & IDML & \textbf{90.6} & \textbf{76.9} & \textbf{42.6} & \textbf{33.8}\\
\hline
\multirow{3}{*}{SOP}
  & Baseline & 79.5 & 91.0 & 53.7 & 50.5\\
  & vMF & 80.8 & 91.6 & 54.0 & 50.8\\
  & IDML & \textbf{81.5} & \textbf{92.3} & \textbf{54.8} & \textbf{51.3}\\
\hline
\end{tabular}
\vspace{-2mm}
\end{table}

\begin{table}[!t] \small
\centering
\caption{Experimental results of out-of-distribution detection. (Tru: Training uncertainty, Teu:Testing uncertainty)}
\vspace{-3pt}
\label{tab:ood}
\setlength\tabcolsep{8.5pt}
\arraysep
\begin{tabular}{lccccc}
\hline
Method & ID & OOD & AUROC & Tru & Teu \\
\hline
Baseline & C-10 & C-100 & 90.6 & - & -\\ 
vMF & C-10 & C-100 & 91.2 & - & -\\ 
IDML & C-10 & C-100 & \textbf{91.9} & 4.56 & 6.91\\
\hline
Baseline & C-100 & C-10 & 55.3 & - & -\\ 
vMF & C-100 & C-10 & 57.6 & - & -\\ 
IDML & C-100 & C-10 & \textbf{58.6} & 5.02 & 7.30\\
\hline
\end{tabular}
\vspace{-2mm}
\end{table}

\begin{table}[!t] \small
\centering
\caption{Comparison experiments with sample weighting methods.}
\label{tab:sample weighting}
\setlength\tabcolsep{7.5pt}
\arraysep
\begin{tabular}{c|cc|cc|cc}
\hline
& \multicolumn{2}{|c}{CUB-200-2011} & \multicolumn{2}{|c}{Cars196} & \multicolumn{2}{|c}{SOP}\\
 \hline
Method & R@1 & NMI & R@1 & NMI & R@1 & NMI\\
\hline
L2RW & 69.7 & 72.6 & 89.4 & 76.2 & 80.7 & 91.6\\ 
MWN & 70.0 & 73.1 & 90.0 & 76.7 & 81.2 & 92.0\\
SPL & 69.4 & 72.6 & 88.7 & 75.5 & 80.2 & 91.2\\
FL & 70.3 & 73.2 & 90.2 & 76.7 & 81.3 & 92.2\\
IDML & \textbf{70.7} & \textbf{73.5} & \textbf{90.6} & \textbf{76.9} & \textbf{81.5} & \textbf{92.3}\\
\hline
\end{tabular}
\tablevspace
\end{table}

\subsubsection{Comparisons with Sampling Weighting Methods}
IDML can be regarded as a sample weighting approach during training. We model the uncertainty of the sample and then use the relative relationship between the uncertainty and the semantic features of the sample to determine the coefficients of the metric function. 
We provide the experimental results in Table~\ref{tab:sample weighting}.
L2RW~\cite{ren2018learning} and  MWN~\cite{shu2019meta} adopt a meta-learning framework mechanism for sample weighting. The focal loss (FL)~\cite{lin2017focal} weights samples with the motivation of hard negative sampling, which improves the efficiency of training.
Differently, self-paced learning (SPL)~\cite{kumar2010self} trains the model with samples from easy to difficult. 
Nevertheless, these two approaches need to pre-specify the weighting function and the hyper-parameters~\cite{shu2019meta}, to which these methods are sensitive.
We observe that IDML outperforms the mentioned methods in our deep metric learning setting (\emph{e.g.} 0.7$\%$ higher than MWN and 0.4$\%$ higher than FL at Recall@1 on the CUB-200-2011 dataset), which demonstrates the superiority of our uncertainty modeling and reweighting strategy.

\begin{table*}[tb] \tablesize
\centering
\caption{Analysis of different components of IDML on the CUB-200-2011, Cars196, and Stanford Online Products datasets.}
\label{tab:component}
\vspace{-3pt}
\setlength\tabcolsep{8.5pt}
\arraysep
\begin{tabular}{l|cccc|cccc|cccc}
\hline
& \multicolumn{4}{|c}{CUB-200-2011} & \multicolumn{4}{|c}{Cars196} & \multicolumn{4}{|c}{Stanford Online Products}\\
 \hline
Method & R@1 & NMI & RP & M@R & R@1 & NMI & RP & M@R & R@1 & NMI & RP & M@R\\
\hline
Margin-DW~\cite{wu2017sampling} & 65.9 & 69.5 & 36.0 & 24.9  & 82.6 & 69.3 & 36.4 & 26.5 & 78.5 & 90.1 & 53.4 & 50.2\\ 
Mixup-MDW & 67.1 & 71.6 & 36.7 & 25.5 & 84.7 & 72.4 & 38.0 & 28.0 & 79.1 & 90.5 & 53.6 & \textbf{50.4}\\
ISM-MDW & 67.0 & 71.4 & 36.9 & 25.7 & 84.4 & 71.9 & 37.9 & 28.1 & 78.9 & 90.4 & 53.6 & 50.3\\
PEL-MDW~\cite{oh2018modeling} & 63.3 & 67.1 & 34.6 & 24.2 & 80.4 & 67.1 & 34.8 & 25.5 & 76.4 & 88.7 & 51.2 & 48.7\\
PEL-Mixup-MDW & 64.5 & 68.6 & 35.3 & 24.7 & 82.3 & 68.9 & 35.9 & 26.1 & 77.2 & 89.4 & 52.1 & 49.3\\
IDML-MDW & \textbf{67.9} & \textbf{72.1} & \textbf{37.2} & \textbf{26.1} & \textbf{86.1} & \textbf{73.0} & \textbf{39.2} & \textbf{29.7} & \textbf{79.4} & \textbf{91.0} & \textbf{53.7} & \textbf{50.4}\\
\hline
ProxyAnchor~\cite{kim2020proxy} & 69.0 & 72.3 & 38.5 & 27.5 & 87.3 & 75.7 & 40.9 & 31.8 & 79.5 & 91.0 & 53.7 & 50.5\\
Mixup-PA & 69.8 & 73.0 & 39.1 & 28.1 & 88.5 & 75.8 & 41.0 & 32.1 & 80.6 & 91.8 & 54.4 & 50.7\\
ISM-PA & 69.5 & 73.1 & 38.9 & 28.0 & 88.8 & 75.8 & 41.2 & 32.2 & 80.3 & 91.8 & 54.3 & 50.9\\
PEL-PA~\cite{oh2018modeling} & 64.9 & 67.1 & 34.5 & 23.7 & 83.4 & 66.4 & 34.4 & 24.9 & 76.8 & 89.7 & 51.8 & 48.7 \\
PEL-Mixup-PA  & 65.7 & 68.0 & 35.6 & 24.7 & 84.5 & 66.6 & 34.6 & 25.1 & 77.9 & 90.5 & 52.6 & 49.9 \\
IDML-PA & \textbf{70.7} & \textbf{73.5} & \textbf{39.3} & \textbf{28.4} & \textbf{90.6} & \textbf{76.9} & \textbf{42.6} & \textbf{33.8} & \textbf{81.5} & \textbf{92.3} & \textbf{54.8} & \textbf{51.3}\\
\hline
\end{tabular}
\vspace{-2mm}
\end{table*}

\begin{table*}[t] \tablesize 
  \centering
  \caption{Experimental results with various forms of uncertain data.}
  \label{tab:uncertainty} 
\vspace{-3pt}
\setlength\tabcolsep{8.3pt}
\arraysep
    \begin{tabular}{l|cccc|cccc|cccc}
    \hline
    \multicolumn{1}{c|}{\multirow{2}[0]{*}{Method}} & \multicolumn{4}{c|}{CUB-200-2011} & \multicolumn{4}{c|}{Cars196} & \multicolumn{4}{c}{Stanford Online Products}  \\ \cline{2-13}
          & R@1   & NMI   & RP    & M@R   & R@1   & NMI   & RP    & M@R  & R@1   & NMI   & RP    & M@R  \\
    \hline
    Baseline & 69.0  & 72.3  & 38.5  & 27.5  & 87.3  & 75.7  & 40.9  & 31.8  & 79.5  & 91.0  & 53.7  & 50.5 \\
    Baseline + ISM & 69.5  & 73.1  & 38.9  & 28.0  & 88.8  & 75.8  & 41.2  & 32.2  & 80.3  & 91.8  & 54.3  & 50.9 \\
    Low-res (200$\times$200) & 67.4  & 71.3  & 37.7  & 26.2  & 87.0  & 75.3  & 40.8  & 31.2  & 78.7  & 90.0  & 52.6  & 49.1 \\
    Low-res + ISM & 68.9  & 71.9  & 38.2  & 27.4  & 89.0  & 76.2  & 41.3  & 32.5  & 79.3  & 90.9  & 53.4  & 50.4 \\
    Blur (p=0.5)  & 69.0  & 72.4  & 38.5  & 27.7  & 88.2  & 75.8  & 41.1  & 32.0  & 79.7  & 91.0  & 53.9  & 50.5 \\
    Blur + ISM & 69.2  & 72.5  & 38.6  & 27.7  & 89.6  & 76.4  & 41.8  & 32.8  & 79.7  & 91.1  & 53.9  & 50.3 \\
    Occlusion (p=0.5) & 69.3  & 72.4  & 38.6  & 27.9  & 87.9  & 75.7  & 41.1  & 31.8  & 80.2  & 91.2  & 54.2  & 50.7 \\
    Occlusion + ISM & 69.6  & 72.8  & 38.8  & 28.0  & 89.2  & 76.4  & 41.5  & 32.6  & 80.6  & 91.7  & 54.6  & 50.9 \\
    Mixup & 69.8  & 73.0  & 39.1  & 28.1  & 88.5  & 75.8  & 41.0  & 32.1  & 80.6  & 91.8  & 54.4  & 50.7 \\
    Mixup + ISM & \textbf{70.7} & \textbf{73.5} & \textbf{39.3} & \textbf{28.4} & \textbf{90.6} & \textbf{76.9} & \textbf{42.6} & \textbf{33.8} & \textbf{81.5} & \textbf{92.3} & \textbf{54.8} & \textbf{51.3} \\
    \hline
    \end{tabular}%
   \vspace{-3mm}
\end{table*}%

\begin{table}[t] \small
  \centering
  \caption{Results with various degrees of data augmentation.}
  \label{tab:chance} 
\vspace{-3pt}
\setlength\tabcolsep{4.5pt}
\arraysep
    \begin{tabular}{l|c|cccc}
    \hline
    Method & Level & R@1 & NMI   & RP    & M@R  \\
    \hline
    Baseline & 5.13  & 87.3  & 75.7  & 40.9  & 31.8  \\
    Baseline + ISM & 5.13  & 88.8  & 75.8  & 41.2  & 32.2  \\
    Low-res (200$\times$200) & 5.59 & 87.0  & 75.3  & 40.8  & 31.2  \\
    Low-res + ISM (200$\times$200) & 5.59 & 89.0  & 76.2  & 41.3  & 32.5  \\
    Low-res (100$\times$100) & 6.98 & 83.2  & 72.6  & 37.3  & 27.5  \\
    Low-res + ISM (100$\times$100) & 6.98 & 84.1  & 73.2  & 37.6  & 27.9  \\
    Blur (p=0.1) & 5.30 & 87.7  & 76.1  & 41.1  & 31.9 \\
    Blur + ISM (p=0.1) & 5.30 & 88.9 & 76.0  & 41.2  & 32.3 \\
    Blur (p=0.5) & 5.99 & 88.2  & 75.8  & 41.1  & 32.0 \\
    Blur + ISM (p=0.5) & 5.99 & 89.6  & 76.4  & 41.8  & 32.8 \\
    Blur (p=1.0) & 6.75 & 86.4 & 74.6 & 40.1  & 30.7 \\
    Blur + ISM (p=1.0) & 6.75 & 87.1 & 75.2 & 40.8  & 31.5 \\
    Occlusion (p=0.1) & 5.24 & 87.3  & 75.3 & 41.0  & 31.9  \\
    Occlusion + ISM (p=0.1) & 5.24 & 88.0  & 75.7 & 41.2  & 32.0  \\
    Occlusion (p=0.5) & 5.62 & 87.9  & 75.7  & 41.1  & 31.8  \\
    Occlusion + ISM (p=0.5) & 5.62 & 89.2  & 76.4  & 41.5  & 32.6  \\
    Occlusion (p=1.0) & 6.21 & 88.2 & 76.0 & 41.2  & 32.0  \\
    Occlusion + ISM (p=1.0) & 6.21 & 89.4 & 76.6  & 41.5  & 32.7  \\
    \hline
    \end{tabular}%
\tablevspace
\end{table}%

\subsubsection{Ablation Study of Different Components}
We conducted experiments with the margin loss and ProxyAnchor loss to analyze the effect of different components of our framework.
Table~\ref{tab:component} shows the experimental results on the CUB-200-2011, Cars196, and Stanford Online Products datasets. 

We first applied Mixup to the baseline method (Mixup-MDW) without using our introspective similarity metric and then only employed the proposed metric without mixup (ISM-MDW).
We see that the Mixup method and the proposed ISM can independently boost the performances of the baseline method.
For example, ISM with the ProxyAnchor loss achieves 69.5$\%$, 88.8$\%$, and 80.3$\%$ at Recall@1 on three datasets, respectively.
Our IDML framework further improves the performance by combining Mixup and our ISM. 
Furthermore, we reproduced the probabilistic embedding learning (PEL) framework~\cite{oh2018modeling} on each loss (PEL-MDW) and also equipped it with Mixup (PEL-Mixup-MDW) for fair comparisons with our framework.
We observe that it achieves lower performance than the baseline method (64.9$\%$, 83.4$\%$, and 76.8$\%$ at Recall@1 with the ProxyAnchor loss on three datasets), and further using Mixup improves the performance.
The performance drop might result from the compromise of discriminativeness when representing images as distributions.
Differently, our framework uses an uncertainty embedding to model the uncertainty which does not affect the discriminativeness of the semantic embedding.

\newcommand\figwidth{0.323}

\begin{figure*}[t]
\centering
\subcaptionbox{ CUB-200-2011\label{subfig:cub}}{
\includegraphics[width=\figwidth\textwidth]{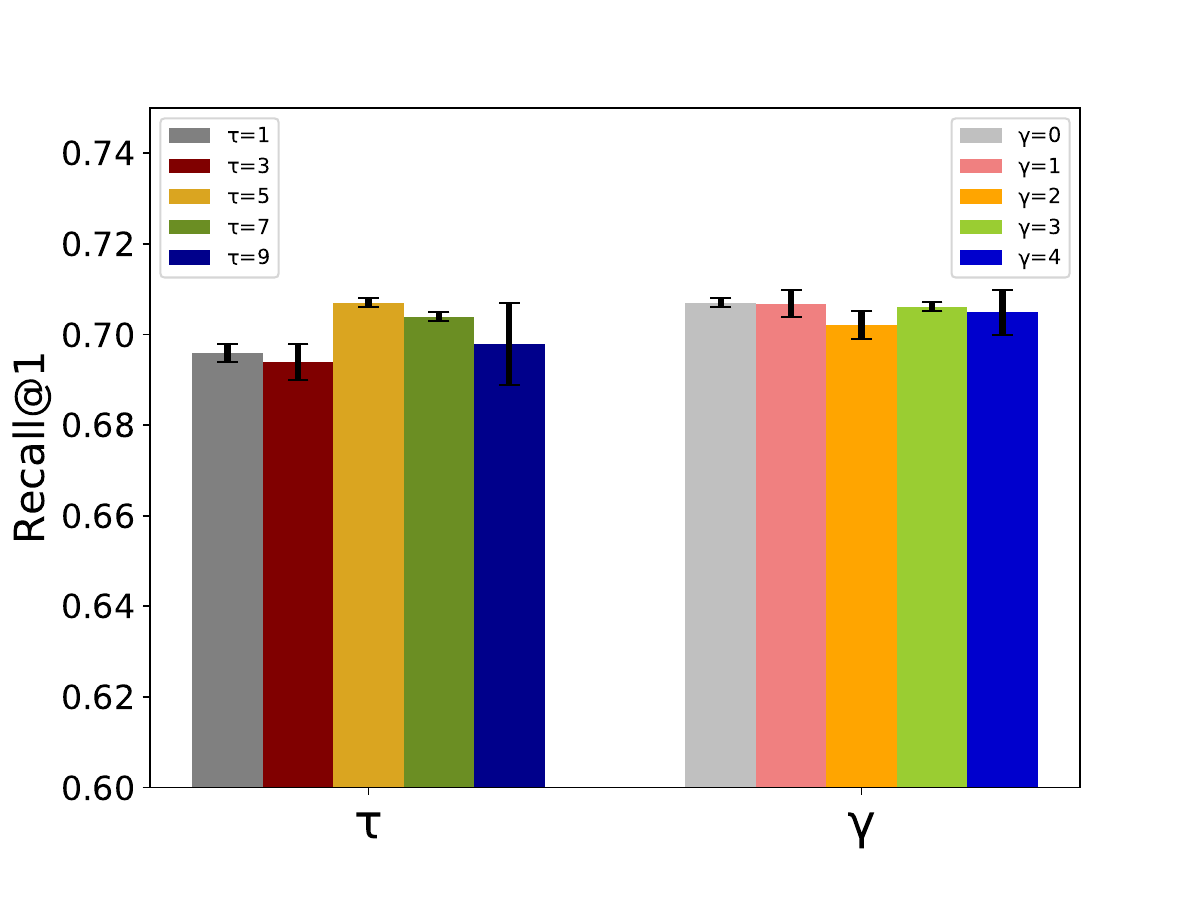}
}\hfill
\subcaptionbox{ Cars196\label{subfig:car}}{
\includegraphics[width=\figwidth\textwidth]{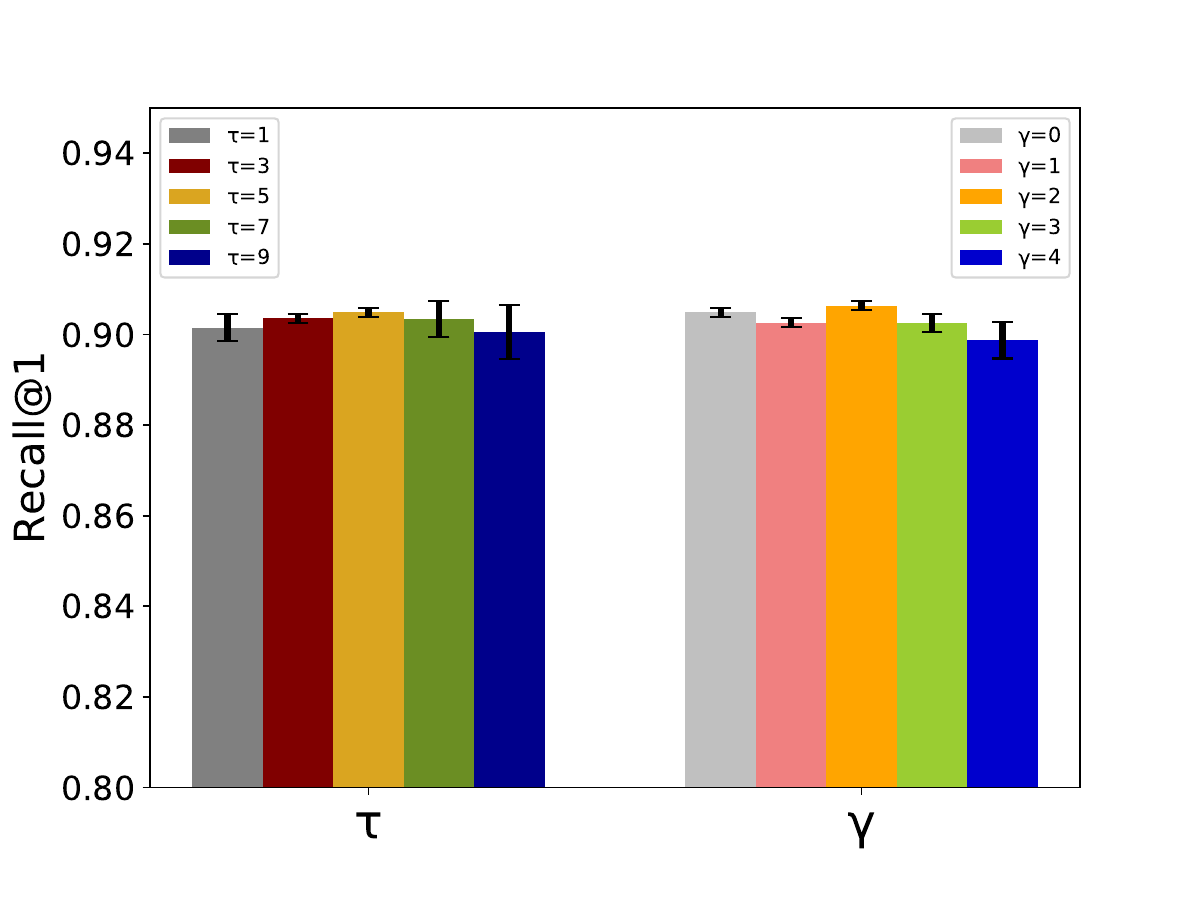}
}\hfill
\subcaptionbox{ 
Stanford Online Products\label{subfig:sop}}{
\includegraphics[width=\figwidth\textwidth]{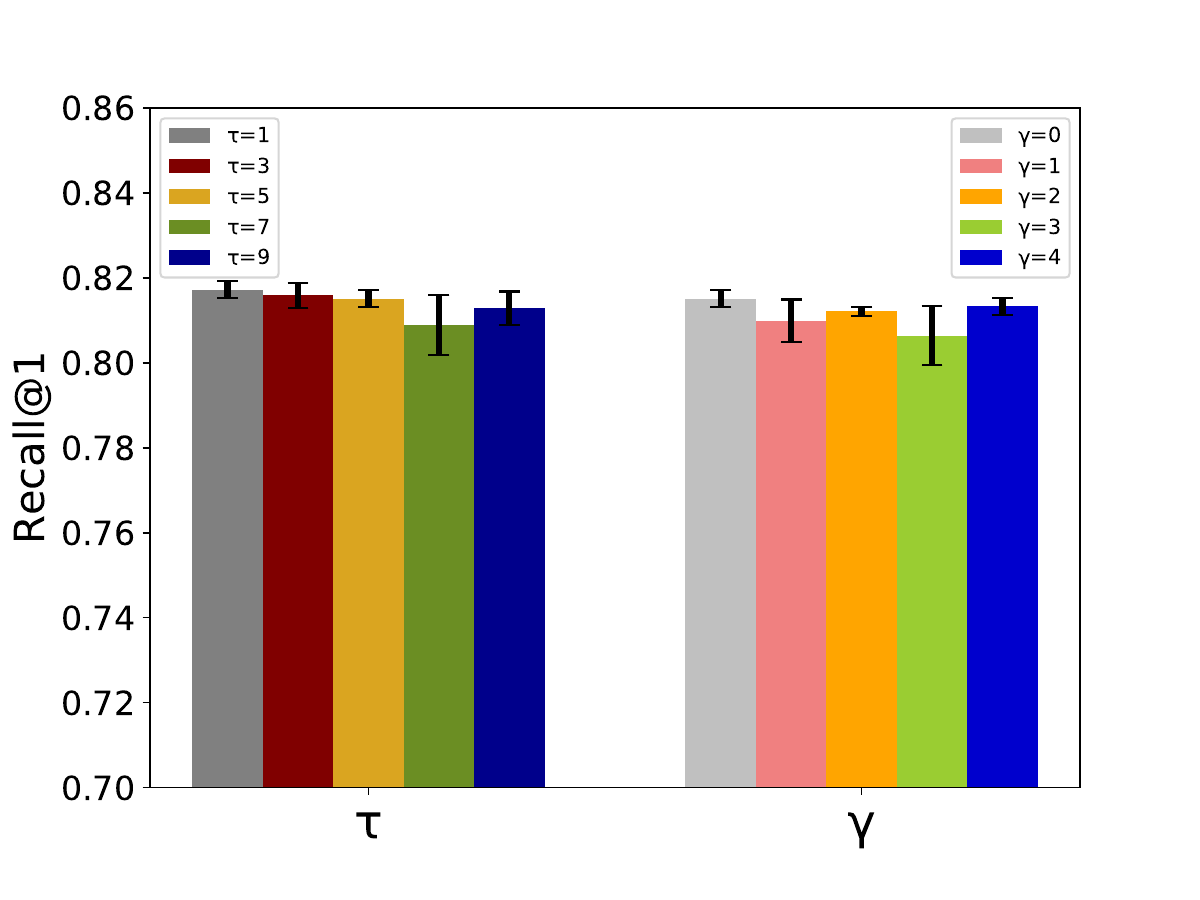}
}
\vspace{-2mm}
\caption{Impact of metric parameters on the CUB-200-2011, Cars196 and Stanford Online Products datasets.}
\label{fig:para}
\figvspace
\end{figure*}

\subsubsection{Uncertainty of Other Forms}
In addition to Mixup, we further conducted experiments when training with lowered-resolution, blurred, and occluded images, as shown in Table~\ref{tab:uncertainty}.
Though certain augmentations (\emph{e.g.}, low-resolution) reduce the performance of the baseline, further applying our ISM consistently attains better results than training without these augmentations.
For example, adopting the low-resolution augmentation achieves 67.4$\%$ at Recall@1 on the CUB-200-211 dataset, which is 1.6$\%$ lower than the baseline, while further applying ISM improves the performance by $1.5\%$ and is close to the original baseline method.
This verifies the effectiveness of our method to deal with various forms of uncertain data. In Table~\ref{tab:chance}, we gradually increased the degree or chance of augmentations.
We adopted a model trained with IDML to output the average uncertainty of images corresponding to each form of augmentation to denote the uncertainty levels and provided the corresponding performances on the Cars196 dataset. We observe that IDML consistently improves the retrieval performance with diverse chances or degrees of adopting data augmentation strategies, demonstrating the generalization of our proposed approach.

\begin{table}[t] \small
\centering
\caption{Analysis of the batch size (BS) on the CUB-200-2011, Cars196, and Stanford Online Products datasets.}
\vspace{-3pt}
\label{tab:bs}
\setlength\tabcolsep{9pt}
\arraysep
\begin{tabular}{c|cc|cc|cc}
\hline
& \multicolumn{2}{|c}{CUB-200-2011} & \multicolumn{2}{|c}{Cars196} & \multicolumn{2}{|c}{SOP}\\
 \hline
BS & R@1 & NMI & R@1 & NMI & R@1 & NMI\\
\hline
40 & 66.3 & 70.6 & 88.2 & 74.1 & 78.1 & 89.5\\ 
60 & 67.1 & 71.3 & 88.9 & 75.3 & 79.9 & 91.3\\
80 & 68.5 & 72.0 & 89.6 & 75.6 & 80.5 & 91.6\\
100 & 69.6 & 72.5 & 90.0 & 76.3 & 81.2 & 92.1\\
120 & \textbf{70.7} & 73.5 & 90.6 & 76.9 & 81.5 & \textbf{92.3}\\
140 & 70.5 & 73.2 & 90.5 & 77.2 & \textbf{81.6} & 92.2\\
160 & \textbf{70.7} & 73.3 & \textbf{90.7} & 76.8 & 81.3 & \textbf{92.3}\\
180 & \textbf{70.7} & \textbf{73.6} & 90.5 & \textbf{77.5} & 81.3 & 92.1\\
\hline
\end{tabular}
 \tablevspace
\end{table}

\subsubsection{Effect of the Hyper-parameters $\gamma$ and $\tau$}
$\gamma$ determines the introspective bias and $\tau$ controls the temperature in our introspective similarity metric. 
They jointly affect the final performance of our framework. 
We experimentally evaluated their impacts on three datasets, as demonstrated in Fig.~\ref{fig:para}. 
The black lines represent the confidence intervals.
We first fixed $\gamma$ to $0$ and set $\tau$ to 1, 3, 5, 7, 9. 
We see that our framework achieves the best recall@1 when $\tau=5$ for the CUB-200-2011 and Cars196 datasets, indicating the favor of a modest weakening degree.
In addition, we fixed $\tau=5$ and set $\gamma$ to 0, 1, 2, 3, 4 for training. 
The experimental results vary on the three datasets.
Specifically, our framework achieves the best performance when $\gamma=0$ on the CUB-200-2011 and Stanford Online Products datasets while $\gamma=2$ on the Cars196 dataset.
This indicates that the metric is more discreet when comparing images on the Cars196 dataset.
Additionally, we observe that the performance becomes worse when $\tau$ and $\gamma$ continue increasing, which demonstrates that a larger temperature $\tau$ and introspective bias $\gamma$ present a disadvantage for similarity computation. Therefore, we suggest controlling the ranges of $\tau$ and $\gamma$ when applying IDML on other datasets.

\begin{table}[t] \small
\centering
\caption{Analysis of the semantic embedding (SE) size on the CUB-200-2011, Cars196, and Stanford Online Products datasets.}
\vspace{-3pt}
\label{tab:semantic_embedding}
\setlength\tabcolsep{7.8pt}
\arraysep
\begin{tabular}{c|cc|cc|cc}
\hline
& \multicolumn{2}{|c}{CUB-200-2011} & \multicolumn{2}{|c}{Cars196} & \multicolumn{2}{|c}{SOP}\\
 \hline
SE Size & R@1 & NMI & R@1 & NMI & R@1 & NMI\\
\hline
32 & 56.9 & 64.0 & 76.8 & 68.0 & 73.0 & 86.2\\ 
64 & 63.1 & 67.2 & 83.1 & 70.6 & 77.3 & 89.1\\
128 & 66.4 & 70.2 & 86.0 & 73.7 & 79.6 & 91.1\\
256 & 68.5 & 71.7 & 89.2 & 76.1 & 80.8 & 92.0\\
512 & 70.7 & 73.5 & \textbf{90.6} & \textbf{76.9} & 81.5 & \textbf{92.3}\\
1024 & \textbf{71.0} & \textbf{73.6} & 90.3 & 76.4 & \textbf{81.7} & \textbf{92.3}\\
\hline
\end{tabular}
\tablevspace
\end{table}

\subsubsection{Effect of the Batch Size}
We conducted experiments on the CUB-200-2011~\cite{wah2011caltech} and Cars196~\cite{krause20133d} datasets to investigate the influence of the batch size during training. 
Specifically, we set the batch size from $40$ to $180$, as shown in Table~\ref{tab:bs}. 
We observe a relatively consistent performance improvement as the batch size increases on both datasets.
This is because larger batch sizes enable richer relation mining among data.
Still, we see that the performance plateaus and even decreases when the batch sizes exceed $120$. Therefore, we set the batch size to 120 for the main experiments for a better balance of performance and computation.

\subsubsection{Effect of Embedding Sizes}
The dimension of the embedding is a crucial factor for the final performance, as verified by several works~\cite{kim2020proxy,zheng2021deepr}. 
During training, the proposed IDML framework simultaneously obtains a semantic embedding and an uncertainty embedding for each image.
During testing, we only use the semantic embeddings for inference, introducing no additional computation cost.
Still, the uncertainty embedding influences the training of the semantic embedding and thus affects the inference performance.

We first fixed the dimension of the uncertainty embeddings to $512$ and used the dimension of 32, 64, 128, 256, 512, and 1024 for the semantic embeddings as shown in Table~\ref{tab:semantic_embedding}. 
We observe that the performance improves as the size of the semantic embedding increases and reaches the top at 512 and 1024. 
We also see that using a dimension of 1024 does not prominently enhance the results, which might result from the information redundancy.

Furthermore, we fixed the semantic embedding size to $512$ and tested the performance using the dimension of 32, 64, 128, 256, 512, and 1024 for the uncertainty embeddings, as shown in Table~\ref{tab:uncertainty-embedding}. 
We see that the performance gradually improves as the uncertainty embedding size increases, and the model achieves the best result when the dimension is $512$.
In summary, we find that using the dimension of 512 for both the semantic embeddings and the uncertainty embeddings achieves the best accuracy/computation trade-off, and we thus adopted them in the main experiments.

\begin{table}[t] \small
\centering
\caption{Analysis of the uncertainty embedding (UE) size on the CUB-200-2011, Cars196, and Standord Online Products datasets.}
\vspace{-3pt}
\label{tab:uncertainty-embedding}
\setlength\tabcolsep{8pt}
\arraysep
\begin{tabular}{c|cc|cc|cc}
\hline
& \multicolumn{2}{|c}{CUB-200-2011} & \multicolumn{2}{|c}{Cars196} & \multicolumn{2}{|c}{SOP}\\
 \hline
UE Size & R@1 & NMI & R@1 & NMI & R@1 & NMI\\
\hline
32 & 69.4 & 72.5 & 89.2 & 75.9 & 80.3 & 91.3\\ 
64 & 69.2 & 72.9 & 89.8 & 76.5 & 80.6 & 91.5\\
128 & 69.5 & 73.3 & 89.8 & 76.7 & 81.1 & 92.0\\
256 & 70.0 & 73.2 & 90.1 & 76.8 & 81.4 & 92.1\\
512 & \textbf{70.7} & \textbf{73.5} & \textbf{90.6} & \textbf{76.9} & \textbf{81.5} & 92.3\\
1024 & 70.4 & 73.4 & 90.4 & 76.7 & 81.4 & \textbf{92.4}\\
\hline
\end{tabular}
\vspace{-1mm}
\end{table}

\begin{table}[t] \small
\centering
\caption{Effect of different uncertainty formulations during training.}
\label{tab:uncertainty representations}
\vspace{-3pt}
\setlength\tabcolsep{5pt}
\arraysep
\begin{tabular}{l|l|cccc}
\hline
Dataset & Training Metric & R@1 & NMI & RP & M@R\\
\hline
\multirow{2}{*}{CUB-200-2011}
  & $||\mathbf{u}_1||_2+||\mathbf{u}_2||_2$ & 70.1 & 73.0 & 38.9 & 28.2\\
  & $||\mathbf{u}_1+\mathbf{u}_2||_2$ & \textbf{70.7} & \textbf{73.5} & \textbf{39.3} & \textbf{28.4}\\
\hline
\multirow{2}{*}{Cars196} 
  & $||\mathbf{u}_1||_2+||\mathbf{u}_2||_2$ & 90.2 & 76.3 & 41.8 & 32.9\\
  & $||\mathbf{u}_1+\mathbf{u}_2||_2$ & \textbf{90.6} & \textbf{76.9} & \textbf{42.6} & \textbf{33.8}\\
\hline
\multirow{2}{*}{SOP}
  & $||\mathbf{u}_1||_2+||\mathbf{u}_2||_2$ & 81.3 & 92.0 & 54.4 & 51.0\\
  & $||\mathbf{u}_1+\mathbf{u}_2||_2$ & \textbf{81.5} & \textbf{92.3} & \textbf{54.8} & \textbf{51.3}\\
\hline
\end{tabular}
\vspace{-5mm}
\end{table}

\begin{figure}[t]
\centering
\includegraphics[width=0.48\textwidth]{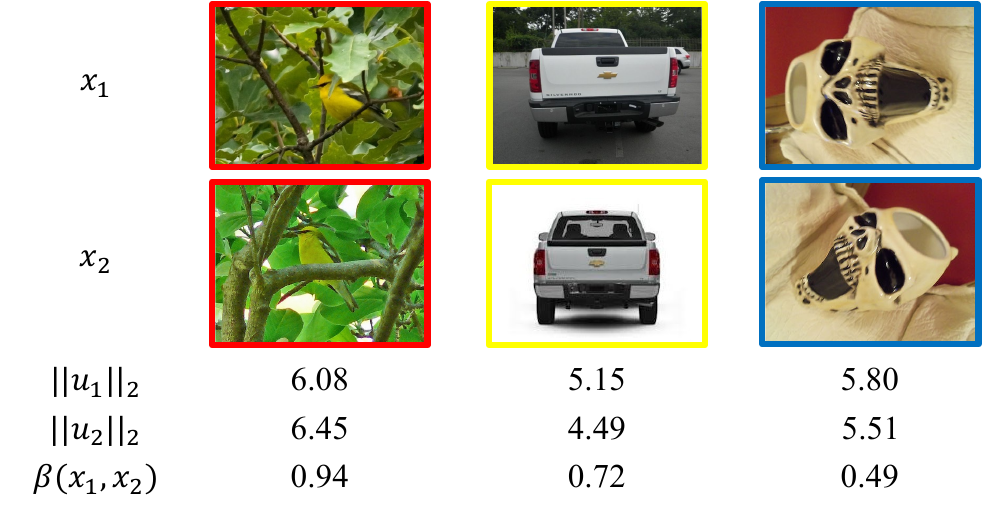}
\vspace{-3mm}
\caption{
Examples of low $\beta(\mathbf{x}_1,\mathbf{x}_2)$ but high $||\mathbf{u}_1||_2$/$||\mathbf{u}_2||_2$.
} 
\label{fig:example}
\vspace{-3mm}
\end{figure}

\begin{table}[t] \small
\centering
\caption{Effect of different metric formulations during training.}
\label{tab:dissimilar}
\vspace{-3pt}
\setlength\tabcolsep{5.2pt}
\arraysep
\begin{tabular}{l|l|cccc}
\hline
Dataset & Training Metric & R@1 & NMI & RP & M@R\\
\hline
\multirow{3}{*}{CUB-200-2011} & Euclidean & 69.0 & 72.3 & 38.5 & 27.5\\ 
  & ISM-Dis~\eqref{eq:dis} & 69.2 & 72.0 & 38.7 & 28.1\\
  & ISM-Sim~\eqref{eq:sim} & \textbf{70.7} & \textbf{73.5} & \textbf{39.3} & \textbf{28.4}\\
\hline
\multirow{3}{*}{Cars196} & Euclidean & 87.3 & 75.7 & 40.9 & 31.8\\
  & ISM-Dis~\eqref{eq:dis} & 89.4 & 75.4 & 41.6 & 32.5\\
  & ISM-Sim~\eqref{eq:sim} & \textbf{90.6} & \textbf{76.9} & \textbf{42.6} & \textbf{33.8}\\
\hline
\multirow{3}{*}{SOP} & Euclidean & 79.5 & 91.0 & 53.7 & 50.5\\
  & ISM-Dis~\eqref{eq:dis} & 80.2 & 91.3 & 54.0 & 50.6\\
  & ISM-Sim~\eqref{eq:sim} & \textbf{81.5} & \textbf{92.3} & \textbf{54.8} & \textbf{51.3}\\
\hline
\end{tabular}
\tablevspace
\end{table}

\begin{table}[t] \small
\centering
\caption{Effect of different metric formulations during testing.}
\label{tab:ISM}
\vspace{-3pt}
\setlength\tabcolsep{3.6pt}
\arraysep
\begin{tabular}{l|l|cccc}
\hline
Dataset & Testing Metric & R@1 & NMI & RP & M@R\\
\hline
\multirow{3}{*}{CUB-200-2011} & Euclidean (baseline) & 69.0 & 72.3 & 38.5 & 27.5\\ 
  & ISM (IDML) & 69.8 & 73.1 & 39.0 & 27.8 \\ 
  & Euclidean (IDML) & \textbf{70.7} & \textbf{73.5} & \textbf{39.3} & \textbf{28.4}\\
\hline
\multirow{3}{*}{Cars196} & Euclidean (baseline) & 87.3 & 75.7 & 40.9 & 31.8\\
  & ISM (IDML) & 89.9 & 76.2 & 42.3 & 33.3\\
  & Euclidean (IDML) & \textbf{90.6} & \textbf{76.9} & \textbf{42.6} & \textbf{33.8}\\
\hline
\multirow{3}{*}{SOP} & Euclidean (baseline) & 79.5 & 91.0 & 53.7 & 50.5\\
  & ISM (IDML) & 80.7 & 92.1 & 54.2 & 50.8\\
  & Euclidean (IDML) & \textbf{81.5} & \textbf{92.3} & \textbf{54.8} & \textbf{51.3}\\
\hline
\end{tabular}
\vspace{-1mm}
\end{table}

\begin{table}[t] \small
\centering
\caption{Effect of the learned uncertainty embedding (UE) against random noise vectors (RNV).}
\label{tab:random vectors}
\vspace{-3pt}
\setlength\tabcolsep{5pt}
\arraysep
\begin{tabular}{l|l|cccc}
\hline
Dataset & UE Formulation & R@1 & NMI & RP & M@R\\
\hline
\multirow{2}{*}{CUB-200-2011}
  & RNV (vMF) & 70.0 & 72.8 & 38.8 & 28.2\\
  & Learned (IDML) & \textbf{70.7} & \textbf{73.5} & \textbf{39.3} & \textbf{28.4}\\
\hline
\multirow{2}{*}{Cars196} 
  & RNF (vMF) & 89.8 & 76.0 & 41.6 & 32.6\\
  & Learned (IDML) & \textbf{90.6} & \textbf{76.9} & \textbf{42.6} & \textbf{33.8}\\
\hline
\multirow{2}{*}{SOP}
  & RNF (vMF) & 81.0 & 91.8 & 54.2 & 50.9\\
  & Learned (IDML) & \textbf{81.5} & \textbf{92.3} & \textbf{54.8} & \textbf{51.3}\\
\hline
\end{tabular}
\vspace{-6mm}
\end{table}

\begin{figure}[t]
\begin{minipage}[t]{0.22\textwidth}
  \centering
\includegraphics[width=1\textwidth]{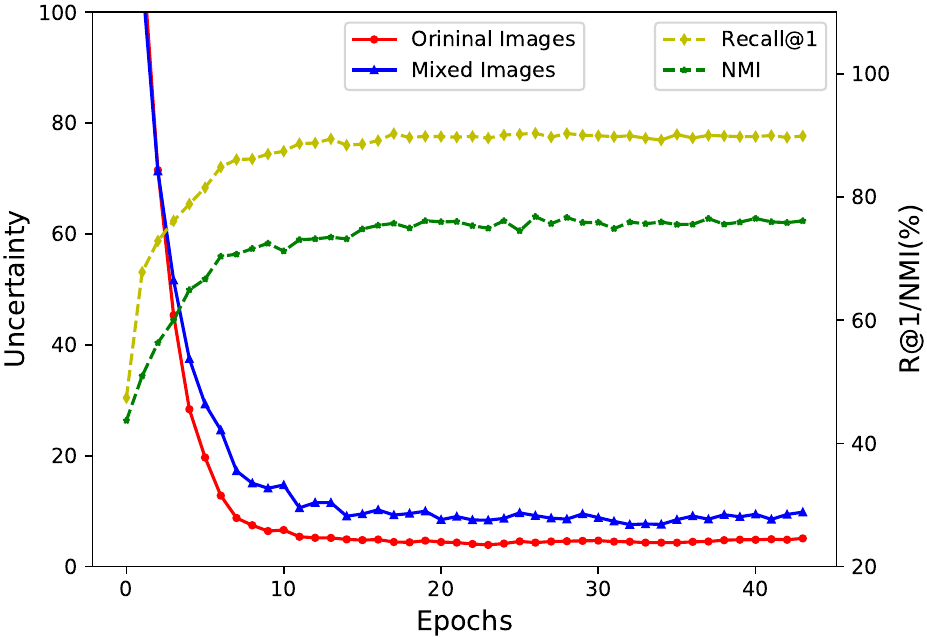}
  \caption{
  The uncertainty trend during training on the Cars196 dataset.
  }
\label{fig:ue}
  \end{minipage}
~~~~
\begin{minipage}[t]{0.22\textwidth}
  \centering
\includegraphics[width=1\textwidth]{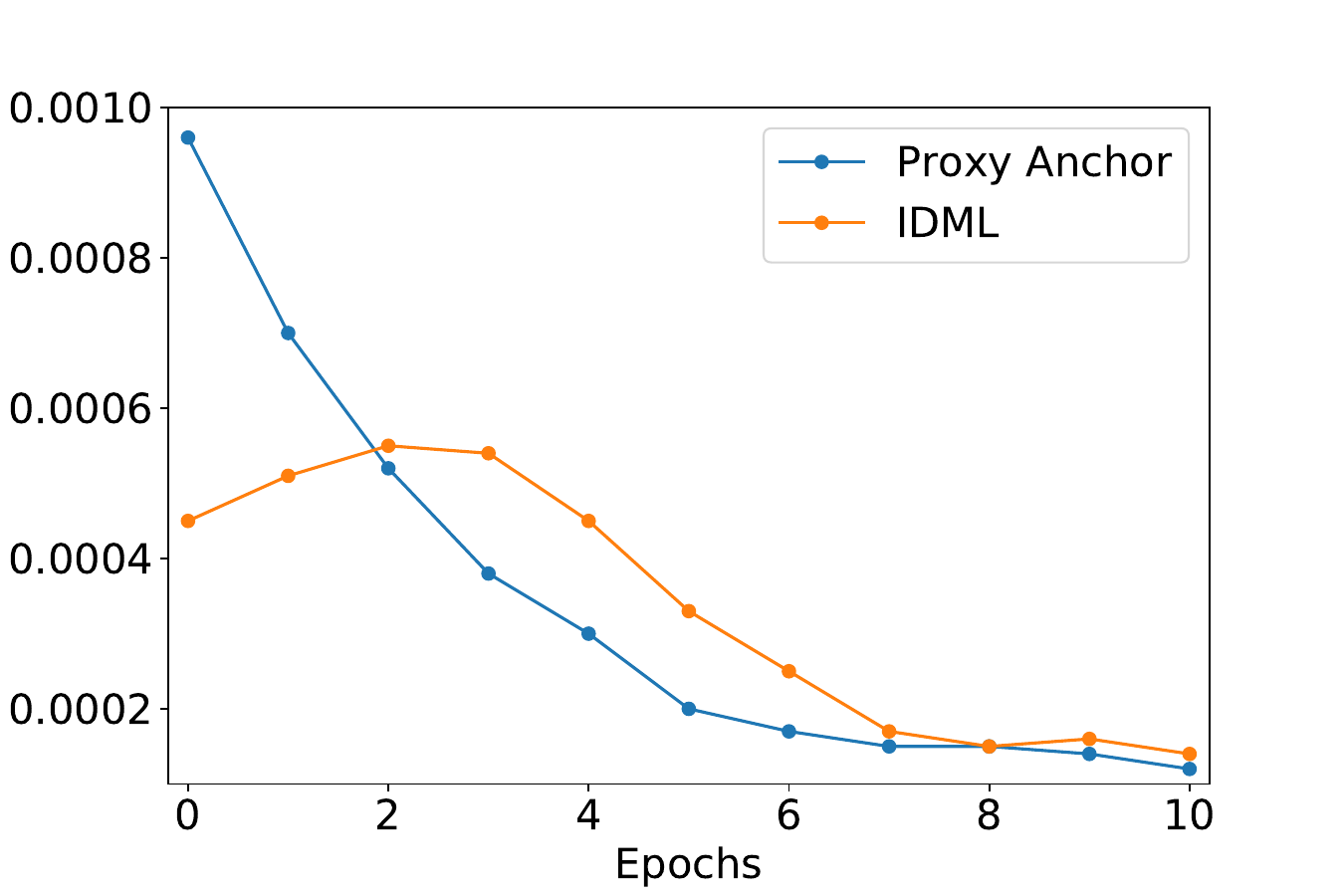}
  \caption{
Gradient changes of IDML and the Proxy Anchor loss during training.
}
\label{fig:gradients}
  \end{minipage}
\vspace{-3mm}
\end{figure}

\begin{table}[t] \small
\centering
\caption{Correlation between relative embeddings.}
\label{tab:correlation}
\vspace{-2mm}
\setlength\tabcolsep{4pt}
\arraysep
\begin{tabular}{l|ccc}
\hline
Metric & Jaccard & MRR & Cosine \\
\hline
Training & 0.0022 $\pm$ 0.0005 & 0.0108 $\pm$ 0.0018 & 0.0076 $\pm$ 0.0009  \\
Testing & 0.0037 $\pm$ 0.0006 & 0.0175 $\pm$ 0.00028 & 0.0113 $\pm$ 0.0015  \\
\hline
\end{tabular}
 \tablevspace
\end{table}

\begin{figure*}[t] 
\centering
\includegraphics[width=0.98\textwidth]{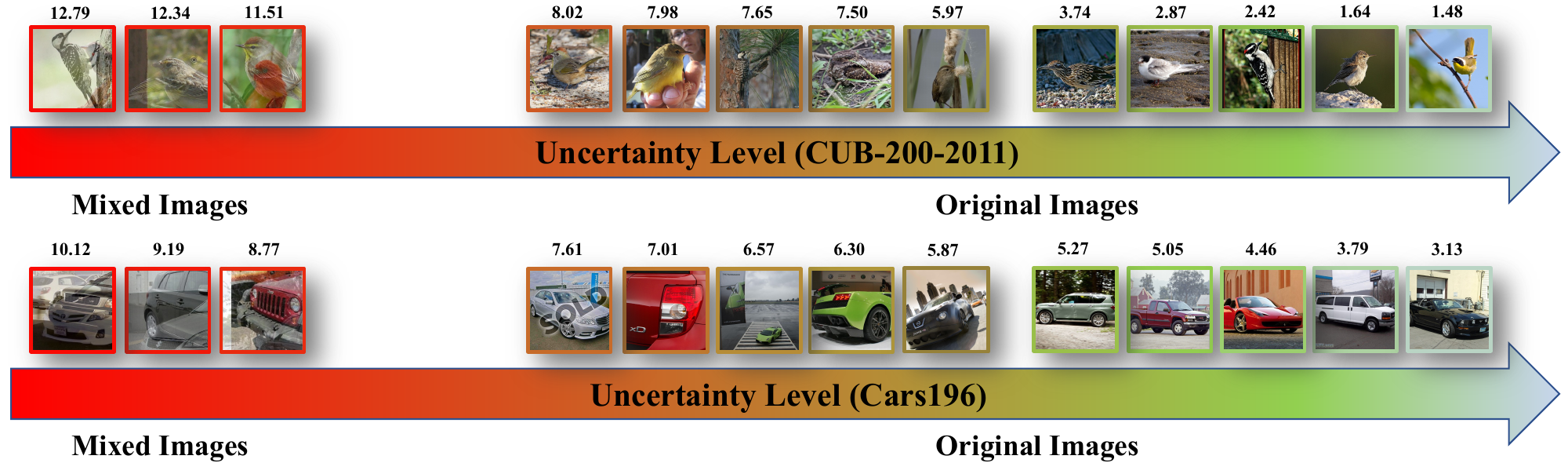}
\caption{Uncertainty levels produced by the proposed IDML framework on the test split of CUB-200-2011 and Cars196.
} 
\label{fig:qr}
\vspace{-3mm}
\end{figure*}

\newcommand\smallfigwidth{0.16}

 \subsubsection{Effect of Different Uncertainty Representations}
 We use $||\mathbf{u}_1+\mathbf{u}_2||_2$ instead of $||\mathbf{u}_1||_2+||\mathbf{u}_2||_2$ for the uncertainty measurement because we consider that the uncertainty computation is related to pairwise samples.
We have also conducted an ablation of these two uncertainty representations, as shown in Table~\ref{tab:uncertainty representations}.
We see that our chosen form of uncertainty computation performs better on the targeted datasets. For example, IDML with $||\mathbf{u}_1||_2+||\mathbf{u}_2||_2$ achieves 70.1$\%$ at Recall@1 on the CUB-200-211 dataset, which is 0.6$\%$ lower than IDML with $||\mathbf{u}_1+\mathbf{u}_2||_2$.
Additionally, we have visualized some examples which have low $\beta(\mathbf{x}_1,\mathbf{x}_2)$ while the uncertainty of each simple is high, presented in Fig.~\ref{fig:example}.
We observe that a sample may have high uncertainty due to background information, diverse angles and vague semantic information. 
However, conditioning the uncertainty measure on both images provides more reasonable results.

\subsubsection{Effect of the Metric Formulation during Training}
When uncertain, the proposed metric tends to treat the pair similarly since we think an uncertain metric should not be able to differentiate all pairs. 
The proposed introspective similarity metric (ISM-Sim) based on the cosine similarity is defined as follows:
\begin{eqnarray}\label{eq:sim}
{C}_{IN}(\mathbf{x}_i, \mathbf{p}_j)=1  -  (1-C(\mathbf{x}_i, \mathbf{p}_j))\cdot e^{(-\frac{1}{\tau} \  \text{r\_conf}(\mathbf{x}_i, \mathbf{p}_j))}.
\end{eqnarray}
Alternatively, we may also weaken the similarity judgment by encouraging the metric to output large distances to all uncertainty pairs.
As a comparison, we additionally modified the metric to treat each ambiguous pair dissimilar (ISM-Dis) as follows:
\begin{eqnarray}\label{eq:dis}
{C}_{IN}(\mathbf{x}_i, \mathbf{p}_j)= C(\mathbf{x}_i, \mathbf{p}_j)\cdot e^{(-\frac{1}{\tau} \  \text{r\_conf}(\mathbf{x}_i, \mathbf{p}_j))}.
\end{eqnarray}
We conducted experiments on the CUB-200-2011 and Cars196 datasets to test the performances of using different metric formulations for training in Table~\ref{tab:dissimilar}. 
We observe that treating each ambiguous pair dissimilar performs worse than the original metric (1.5$\%$, 1.2$\%$, and 1.3$\%$ lower at Recall@1 on three datasets, respectively).
This verifies our motivation for using uncertainty to weaken the semantic discrepancy.

\subsubsection{Effect of the Metric Formulation during Testing}
During testing, we adopt the original similarity metric without our uncertainty-aware term. 
As an alternative, we conducted an experiment using the introspective similarity metric (ISM) during testing on the CUB-200-2011, Cars196, and Stanford Online Products datasets, as shown in Table~\ref{tab:ISM}. 
We observe a decrease in performance when using ISM during testing (0.9$\%$, 0.7$\%$, and 0.8$\%$ lower at Recall@1 on three datasets, respectively), indicating a harmful effect of using uncertainty to weaken the similarity discrepancy during inference.
This is reasonable since providing a clear similarity judgment is more beneficial to discriminative tasks such as image retrieval.

\subsubsection{Effect of the Learned Uncertainty Embeddings against Random Noise Vectors}
We provided comparison experiments when the uncertainty embeddings are not learned but randomly sampled from a vMF~\cite{scott2021mises} around each embedding. 
The comparison experimental results are presented in Table~\ref{tab:random vectors}. 
We observe that the performance becomes worse when the uncertainty embeddings are randomly sampled (0.7$\%$, 0.8$\%$, and 0.5$\%$ lower at Recall@1 on three datasets, respectively). This demonstrates that the performance improvements of IDML indeed stem from the actual aspect of learning uncertainty-aware representations and not just random noise injection into the training process.

\newcommand\basefigwidth{0.498}
\begin{figure*}[t]
    \centering
    \begin{subfigure}[t]{\basefigwidth\textwidth}
        \centering
        \includegraphics[width=\textwidth]{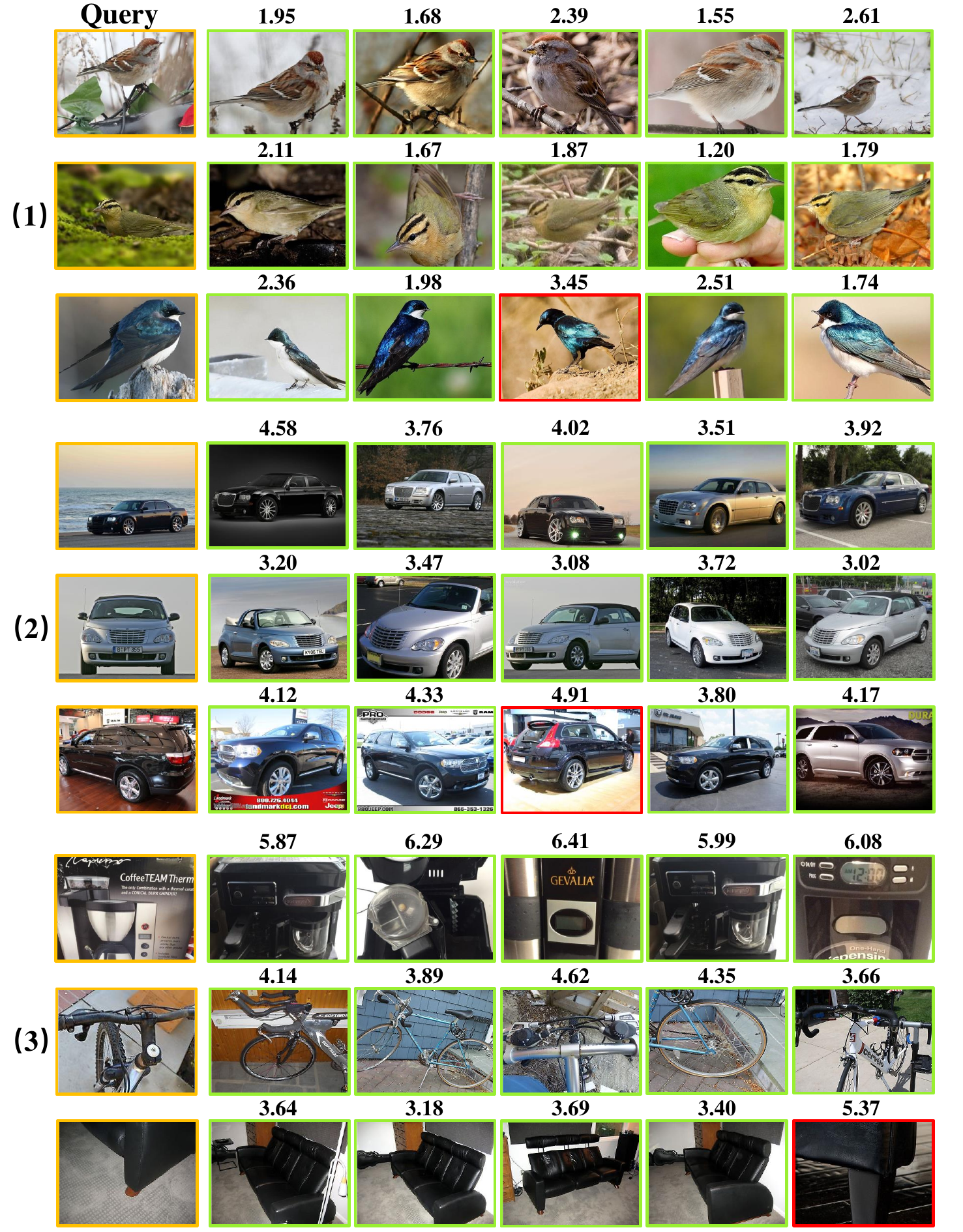} 
        \caption{IDML}
        \label{IDML}
    \end{subfigure}%
    \begin{subfigure}[t]{\basefigwidth\textwidth}
        \centering
        \includegraphics[width=\textwidth]{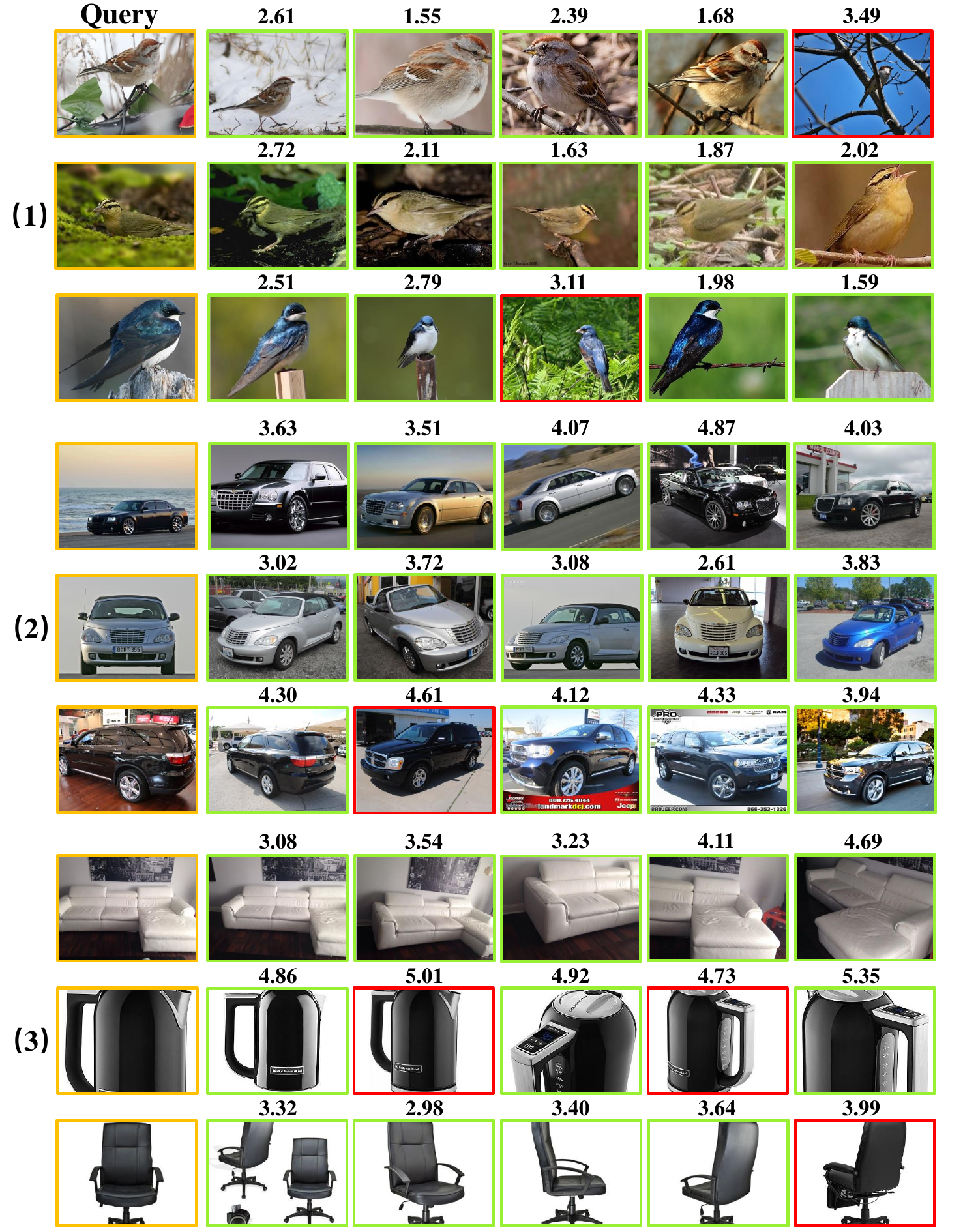}
        \caption{Proxy Anchor}
        \label{Proxy Anchor}
    \end{subfigure}
\vspace{-1mm}
    \caption{Visualization of the top-5 retrieved samples of our IDML framework on (1) CUB-200-2011, (2) Cars196, and (3) Stanford Online Products datasets. 
We use the orange, green, and red color to denote the query, positive, and negative sample, respectively.}
    \label{fig:visualization}
\vspace{-5mm}
\end{figure*}

\subsubsection{Uncertainty Trend during Training}
To demonstrate that our IDML framework properly handles mixed images with high uncertainty, we visualize the trend of uncertainty level for both original images and mixed images during training, as shown in Fig.~\ref{fig:ue}.
We define the uncertainty level of an image to be the L2-norm of its uncertainty embedding.
We see that the uncertainty decreases for both original and mixed images as the training proceeds and becomes stable as the model converges. 
We also observe that the uncertainty level for mixed images is larger than that of original images.
This verifies that our framework can indeed learn the uncertainty in images.

\subsubsection{Comparisons of Gradient Changes between IDML and the Baseline}
We provide the comparison of the average gradient changes of the semantic embedding between IDML and the Proxy Anchor loss during training, presented in Fig.~\ref{fig:gradients}.
We observe that the gradient of IDML first increases and then gradually decreases.
That is because IDML outputs relatively higher uncertainties for samples at the beginning of training, which weakens the corresponding gradient of the semantic embedding. 
As the uncertainty decreases, the gradient of the model rises to a certain extent.
However, in the subsequent training process, as the prediction result of IDML approaches the ground truth, the corresponding gradient gradually decreases, similar to the baseline Proxy Anchor loss.
Therefore, our IDML learns at an adaptive and slower pace to deal with the uncertainty during training.

\subsubsection{Correlation between Semantic Embeddings and Uncertainty Embeddings}
We converted the absolute embedding into the relative embedding and then calculated the correlation between samples according to~\cite{moschella2022relative}.
Specifically, we adopted the uniform sampling strategy, which samples the anchors with a uniform probability distribution over all the available samples. Then, we calculated the cosine similarity between each sample and the anchors in the embedding space followed by the formulation of the relative embedding, denoted as:
\begin{equation}
r_{\mathbf{x}_i}=(C(\mathbf{x}_i,\mathbf{a}_1),C(\mathbf{x}_i,\mathbf{a}_2),...,C(\mathbf{x}_i,\mathbf{a}_K)),
\end{equation}
where $K$ denotes the number of anchors (K=100 in our experiment), $\mathbf{a}_i$ denotes the embedding of the $i$th anchor, $C(\cdot,\cdot)$ denotes the cosine similarity between two embeddings. 

In the proposed IDML framework, we acquired the relative semantic embeddings and the relative uncertainty embeddings according to the above process. Under such circumstances, we obtained the correlation between the relative semantic embeddings and the relative uncertainty embeddings by calculating the discrete Jaccard similarity (Jaccard), Mean Reciprocal Rank (MRR), and the cosine similarity (Cosine) between them, similar to~\cite{moschella2022relative}. The experimental results are illustrated in Table~\ref{tab:correlation}. We observe a consistent low correlation between the relative semantic embeddings and the relative uncertainty embeddings, which demonstrates that the semantic embeddings and uncertainty embeddings are quite disentangled under our IDML framework.

\subsubsection{Uncertainty Levels on the Test Split}
We visualize the uncertainty levels on the test split of the CUB-200-2011 and Cars196 datasets, as shown in Fig.~\ref{fig:qr}. 
We obtain the uncertainty levels of the mixed images together with original images in the test set after the model converges. 
We observe that the uncertainty of mixed images is much larger than that of the original test images since the mixed images contain the information of two images.
Also, we see that several original images result in relatively higher uncertainty than others because of the natural noise such as occlusion and improper directions. 
This further verifies that the proposed framework can successfully learn the uncertainty in images.

\subsubsection{Qualitative Results}
We provide a visualization of several retrieved examples of the proposed IDML framework and ProxyAnchor loss from the CUB-200-2011, Cars196, and Stanford Online Products datasets in Fig.~\ref{fig:visualization}. 
We marked the uncertainty number on the top of each retrieved sample.
The main challenge is the various backgrounds, poses, and viewpoints for the same class on CUB-200-2011, Cars196, and Stanford Online Products, respectively.
Still, our framework can successfully identify the positive samples despite the large intraclass variations.
Additionally, the failure cases only possess subtle differences with the query, which are difficult even for humans to capture. On the contrary, we observe that samples with higher uncertainties appear when not using IDML.

\section{Conclusion}
In this paper, we have presented an introspective deep metric learning framework to more effectively process the semantic uncertainty in training data for better performance. 
We represent an image with a semantic embedding and an uncertainty embedding to model the semantic characteristics and the uncertainty, respectively.
We have further proposed an introspective similarity metric to compute an uncertainty-aware similarity score, which weakens semantic discrepancies for uncertain images. 
We have performed various experiments on three benchmark datasets on image retrieval to analyze the effectiveness of our framework. 
Experimental results have demonstrated a constant performance boost to various methods in different settings.

\ifCLASSOPTIONcompsoc
  \section*{Acknowledgments}
\else
  \section*{Acknowledgment}
\fi

This work was supported in part by the National Key Research and Development Program of China under Grant 2022ZD0114903 and in part by the National Natural Science Foundation of China under Grant 62125603.

\ifCLASSOPTIONcaptionsoff
  \newpage
\fi

{\small
% Generated by IEEEtranS.bst, version: 1.12 (2007/01/11)

}

\newcommand{\controlspace}{\vspace{-10pt}} %

\vspace{-20pt}
\begin{IEEEbiography}[{\includegraphics[width=1in,height=1.25in,clip,keepaspectratio]{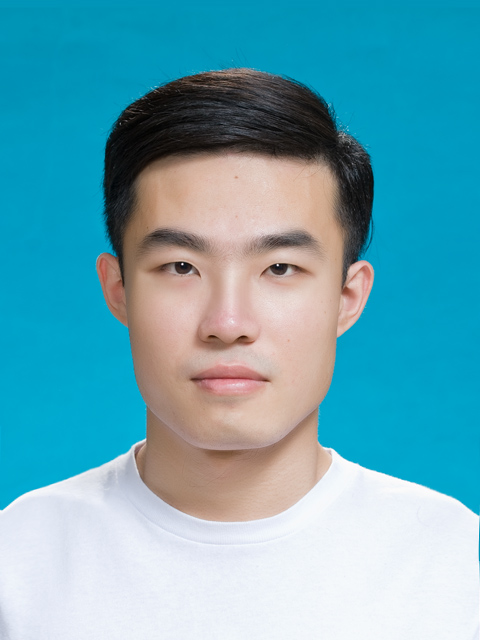}}]{Chengkun Wang}
received the B.S. degree in the Department of Electrical Engineering, Tsinghua University, China, in 2020. He is currently a Ph.D. Candidate with the Department of Automation, Tsinghua University, China. His current research interests include computer vision, metric learning, and representation learning. 
He has authored conference papers in CVPR.
He serves as a regular reviewer member for a number of journals and conferences, e.g. CVPR, TIP, ICME, and ICIP. 
\end{IEEEbiography}

\vfill
\controlspace

\begin{IEEEbiography}[{\includegraphics[width=1in,height=1.25in,clip,keepaspectratio]{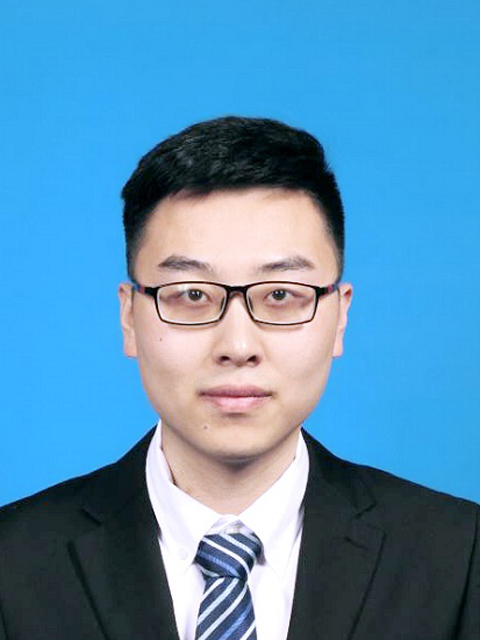}}]{Wenzhao Zheng}
received the B.S. degree in the Department of Physics, Tsinghua University, China, in 2018. 
He is currently a Ph.D. Candidate with the Department of Automation, Tsinghua University, China. His current research interests include omni-supervised representation learning, vision-centric autonomous driving, and explainable artificial intelligence. 
He has authored more than 20 papers in TPAMI, TIP, CVPR, ICCV, ECCV, and ICLR.
He serves as a regular reviewer member for a number of journals and conferences, e.g. TPAMI, TIP, TBIOM, CVPR, ICCV, ECCV, NeurIPS, AAAI, IJCAI, and ICME. 
He is a student member of the IEEE.
\end{IEEEbiography}

\vfill
\controlspace

\begin{IEEEbiography}
[{\includegraphics[width=1in,height=1.25in,clip,keepaspectratio]{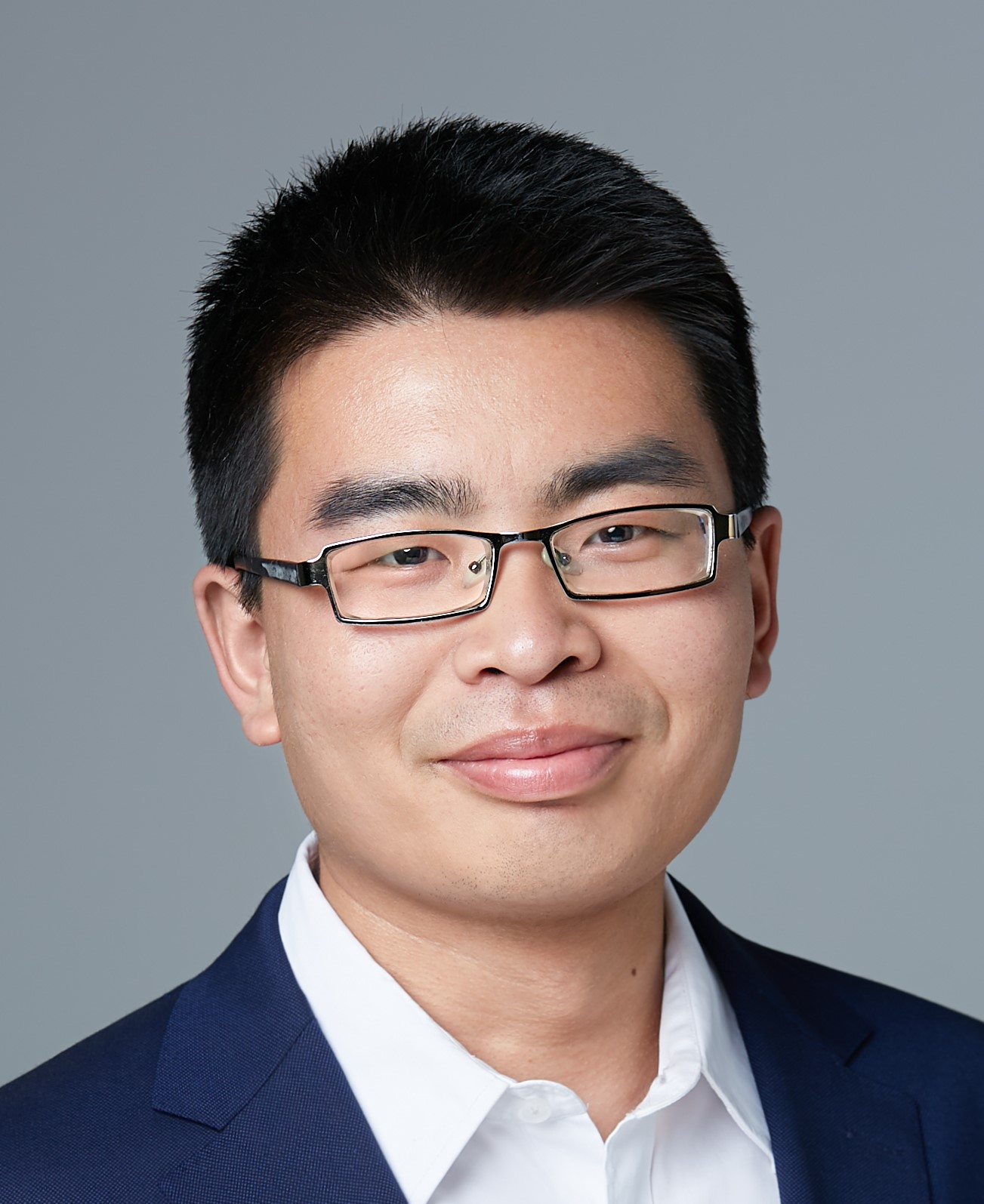}}]{Zheng Zhu} is currently a postdoctoral fellow at Tsinghua University. Before that, he received Ph.D. degree from Institute of Automation, Chinese Academy of Sciences in 2019.  He served reviewers in various journals and conferences including TPAMI, TIP, TMM, TCSVT, CVPR, ICCV, ECCV, ICLR. He has co-authored more than 40 journal and conference papers mainly on computer vision and robotics problems, such as face recognition, visual tracking, human pose estimation, and servo control. He has more than 3,000 Google Scholar citations to his work. He organized the Masked Face Recognition Challenge \& Workshop in ICCV 2021. He ranked the 1st on NIST-FRVT Masked Face Recognition, won the COCO Keypoint Detection Challenge in ECCV 2020 and Visual Object Tracking (VOT) Real-Time Challenge in ECCV 2018. He is a member of the IEEE.
\end{IEEEbiography}

\vfill
\controlspace

\begin{IEEEbiography}[{\includegraphics[width=1in,height=1.25in,clip,keepaspectratio]{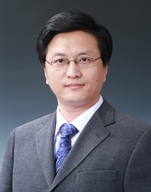}}]{Jie Zhou}
(M'01-SM'04) received the BS and MS degrees both from the Department of Mathematics, Nankai University, China, in 1990 and 1992, respectively, and the PhD degree from the Institute of Pattern Recognition and Artificial Intelligence, Huazhong University of Science and Technology (HUST), China, in 1995. From then to 1997, he served as a postdoctoral fellow in the Department of Automation, Tsinghua University, China. Since 2003, he has been a full professor in the Department of Automation, Tsinghua University. His research interests include computer vision and pattern recognition. In recent years, he has authored more than 100 papers have been published in TPAMI, TIP and CVPR. He is an associate editor for TPAMI and two other journals. He received the National Outstanding Youth Foundation of China Award. He is a senior member of the IEEE and Fellow of the IAPR.
\end{IEEEbiography}

\vfill
\controlspace

\begin{IEEEbiography}[{\includegraphics[width=1in,height=1.25in,clip,keepaspectratio]{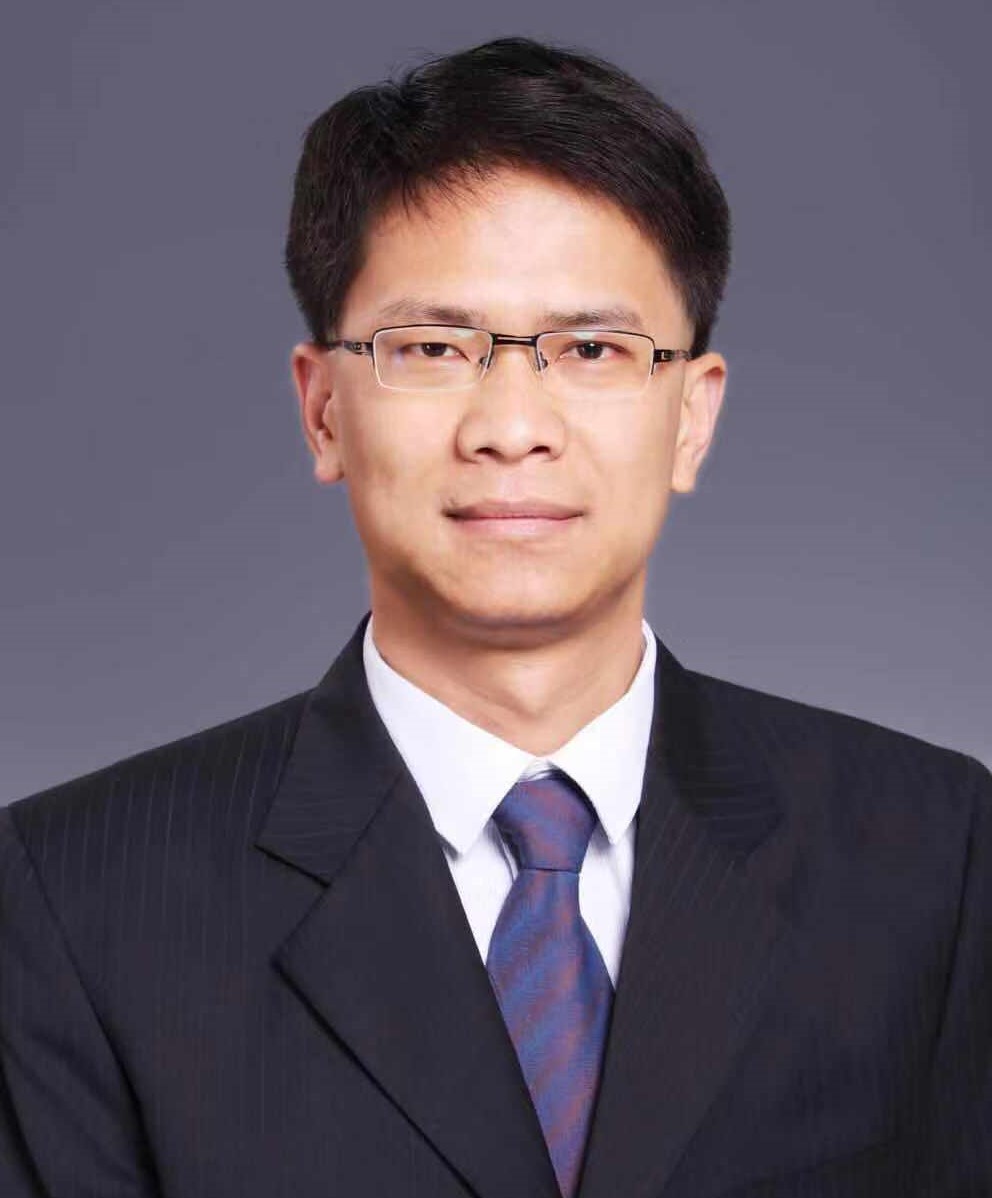}}]{Jiwen Lu}
(M'11-SM'15) received the B.Eng. degree in mechanical engineering and the M.Eng. degree in electrical engineering from the Xi'an University of Technology, China, in 2003 and 2006, respectively, and the Ph.D. degree in electrical engineering from Nanyang Technological University, Singapore, in 2012. He is currently an Associate Professor with the Department of Automation, Tsinghua University, China. His current research interests include computer vision and pattern recognition. He was/is a member of the Multimedia Signal Processing Technical Committee and the Information Forensics and Security Technical Committee of the IEEE Signal Processing Society, and a member of the Multimedia Systems and Applications Technical Committee and the Visual Signal Processing and Communications Technical Committee of the IEEE Circuits and Systems Society. He serves as the Co-Editor-of-Chief for PRL, an Associate Editor for the TIP, TCSVT, the TBIOM, and Pattern Recognition. He also serves as the Program Co-Chair of IEEE FG’2023, VCIP’2022, AVSS’2021 and ICME’2020. He received the National Outstanding Youth Foundation of China Award. He is an IAPR Fellow.

\end{IEEEbiography}

\vfill

\end{document}